%% file: main.tex
\documentclass[10pt]{article} 

\usepackage[accepted]{rlj} 

\usepackage{amssymb}            
\usepackage{mathtools}          
\usepackage{mathrsfs}           
\usepackage{graphicx}           
\usepackage{subcaption}         
\usepackage[space]{grffile}     
\usepackage{url}                
\usepackage{lipsum}             

\usepackage{amsthm}
\theoremstyle{plain}
\newtheorem{theorem}{Theorem}[section]

\theoremstyle{definition}

\theoremstyle{remark}
\newtheorem{remark}[theorem]{Remark}
\usepackage[textsize=tiny, disable]{todonotes}

\definecolor{ourorange}{HTML}{B2C4DB}
\definecolor{ourblue}{rgb}{0.2982297551789076, 0.4433145009416194, 0.6528813559322034}
\definecolor{ourgreen}{rgb}{0.572, 0.586, 0.0}
\definecolor{ourdarkred}{rgb}{1., 0., 0.}

\usepackage{dashrule}
\usepackage{tikz}
\usetikzlibrary{decorations.markings}

\usepackage{lipsum}

\input{al_group_header}

\usepackage{booktabs}
\usepackage[capitalize]{cleveref}

\input{math_commands.tex}

\usepackage{wrapfig}
\usepackage{caption}
\usepackage{subcaption}
\usepackage{graphicx}
\usepackage{url}
\hypersetup{
	colorlinks=true,       
	linkcolor=ourdarkblue,        
	citecolor=ourdarkgreen,        
	filecolor=ourdarkblue,     
	urlcolor=ourviolet
}

\newif\ifcomments
\commentstrue 

\def\piz{{\pi_z}}

\newcommand{\method}{\algo{FBEE$^Q$}\xspace}

\def\qvert{{\,\vert\,}}

\title{Epistemically-guided forward-backward exploration}

\setrunningtitle{Epistemically-guided forward-backward exploration}

\author{Núria Armengol Urpí\textsuperscript{1}, Marin Vlastelica\textsuperscript{1}, Georg Martius\textsuperscript{2}, Stelian Coros\textsuperscript{1}}

\emails{nuriaa@ethz.ch}

\affiliations{
$^{1}$\textbf{Department of Computer Science, ETH Zurich}\\
$^{2}$\textbf{Department of Computer Science, University of Tübingen}\\
}

\contribution{
    This paper phrases the exploration problem for zero-shot RL as uncertainty minimization of a posterior over occupancy measures for a particular representation of an occupancy measure.
    The main difference to previous work is that, while previous work considers completely off-policy exploration algorithms to collect data, this paper considers the uncertainty of the model for data collection in an unsupervised RL setting.
    }
    {
    The representation for occupancy measure used is the \algo{FB}-representation~\citep{touati2021learning} which encodes all optimal policies.  We use an ensemble method approximation to the posterior distribution. 
    Crucially, because of non-uniqueness, the \algo{FB} representation does not allow simple modeling of the posterior uncertainty over \algo{FB} via ensemble disagreement -- there is a necessity of having a single  $B$ representation in order to have an informative notion of uncertainty.
    Furthermore, the $F$-uncertainty is projected to the more practical uncertainty over $Q$-functions for particular latent policy conditioning $z$. 
    }

\contribution{
    We introduce an efficient algorithm for exploration tailored to forward-backward (FB) representations which can be seen as a variant of \emph{uncertainty sampling}~\citep{lewis94}.
    }
    {
    The algorithm relies on sampling a posterior-mean greedy policy $\piz$ which has highest uncertainty in the predictive posterior distribution for a particular state $s$ and executing it in the environment. This exploration strategy, while simple and not considering correlation in uncertainty reduction across all policies $\piz, z \in \gZ$, is a surprisingly efficient method for exploration in \algo{FB} representations. 
    }

\contribution{
    Experimental validation of proposed exploration on several continuous control environments from the DeepMind Control suite~\citep{tassa2018deepmind} in the online learning setting, where we evaluate zero-shot performance on different reward functions within several environments.
    }{
     There is no notion of exploitation in the unsupervised RL setting, therefore there is no need to balance the exploration-exploitation trade-off when collecting data.
     This setup is fundamentally different than single-task online learning, where typically we balance an intrinsic exploration signal or noise with the extrinsic task reward.
     }


\summary{
The goal of zero-shot RL is to provide algorithms for recovering optimal policies for all possible 
reward functions given interaction data with the environment.
Naturally, how well we can recover the optimal policies highly depends on the quality of the data used
to learn them.
Up until now, most algorithms leverage decoupled exploration policies for collecting data in order to learn 
a generalist representation of all optimal policies.
A central argument to this paper is that the exploration policy should not be completely decoupled from
the zero-shot algorithm and should try to minimize the uncertainty that the algorithm has of its representations.
We frame the exploration problem for zero-shot RL as minimization of the epistemic uncertainty on the learned value functions, and realize this in the case of well familiar algorithm, forward-backward (FB) representations.
Crucially, in several empirical settings, using an exploration policy that maximizes the cumulative epistemic uncertainty 
of the FB representation leads to significant improvements of the algorithm's sample complexity.
This enables us to learn well-performing policies fast, with fewer amount of data than other exploration approaches.
}

\begin{document}

\makeCover  
\maketitle  

\begin{abstract}
Zero-shot reinforcement learning is necessary for extracting optimal policies in absence of concrete rewards for fast adaptation to future problem settings.
Forward-backward representations (\algo{FB}) have emerged as a promising method for learning optimal policies in absence of rewards via a factorization of the policy occupancy measure.
However, up until now, \algo{FB} and many similar zero-shot reinforcement learning algorithms have been decoupled from the exploration problem, generally relying on other exploration algorithms for data collection.
We argue that \algo{FB} representations should fundamentally be used for exploration in order to learn more efficiently. 
With this goal in mind, we design exploration policies that arise naturally from the \algo{FB} representation that minimize the posterior variance of the \algo{FB} representation, hence minimizing its epistemic uncertainty.
We empirically demonstrate that such principled exploration strategies improve sample complexity of the \algo{FB} algorithm considerably in comparison to other exploration methods. Code is available at \url{https://sites.google.com/view/fbee-url}.
\end{abstract}

\begin{figure}[htb]
    \centering
    \includegraphics[width=0.85\linewidth]{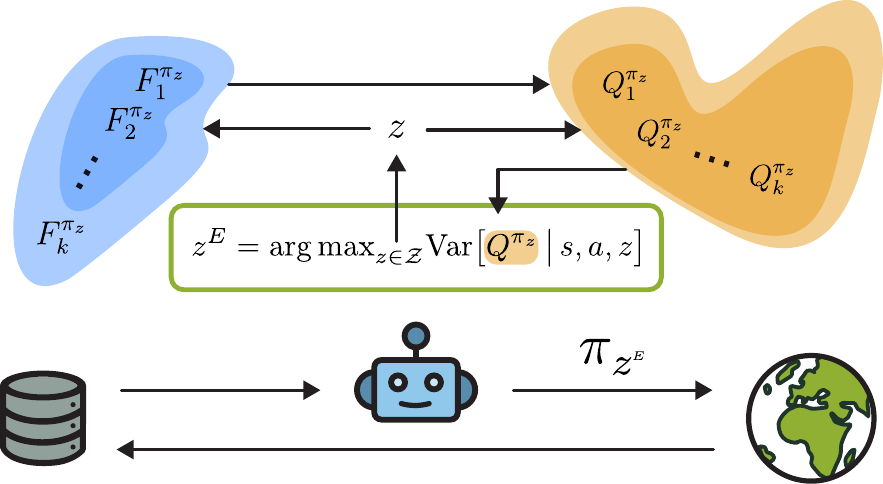}
    \caption{We condition an exploration policy on a reward embedding $z \in \gZ$ maximizing the predictive variance of $Q^\piz$, and execute it for collecting data during learning.
    At inference time, we compute the reward embedding $z$ based on reward evaluation of the dataset, aligned with \citet{touati2021learning}.}
    \label{fig:schematic}
\end{figure}

\section{Introduction}

Reinforcement learning (RL) provides a framework to obtain optimal or near-optimal policies 
from sub-optimal data given a reward function.
However, we cannot possibly enumerate all rewards which are of interest to solve in the future, and hence most RL approaches rely on fixed rewards for training, limiting the generalizability of the learned policies to new tasks.
Zero-shot RL aims to close this gap, by learning optimal policies for all possible reward functions.
In this way, an agent may, with a minimal amount of extra computation, infer an optimal policy for any reward function given at test time. \looseness -1

There are several zero-shot RL methods that have been proposed to  solve this problem.
The earliest instantiation of such methods is that of the successor representation (SR) in the tabular setting~\citep{dayan1993improving},
which has subsequently been extended to the continuous setting with function approximations ~\citep{barreto2017successor}.
The caveat of \algo{SR} is the need to assume a linear dependence between the reward and a feature map, which needs to be handcrafted in advance by the user. This approach cannot easily tackle generic rewards or goal-oriented RL. In the goal-oriented setting, for example, it would require introducing one feature per possible goal state, requiring infinitely many features in continuous spaces. Several frameworks have been proposed to learn this feature map efficiently~\citep{hansen2019fast,liu2021aps,wu2018laplacian}.
More recent work has proposed forward-backward (\algo{FB}) representations~\citep{touati2021learning}, which aims to factorize the 
occupancy distribution of the policies into a forward representation ($F$) of the current state and backward representation ($B$) of a target state.
While the linearity of \algo{SR}'s allows us to infer the optimal policy by solving linear regression onto the sampled rewards,
\algo{FB} infers an optimal policy by Monte Carlo estimation of an integral, which, given a well-learned factorization of the 
occupancy distribution yields the optimal policy representation $z$ for any given reward function.
A critical part in both \algo{FB} and \algo{SR} frameworks is that of learning an accurate occupancy distribution (or successor measure ) for all policies, which requires observing significant amount of environment state transitions.

Up until now, this problem has been tackled by using exploration policies that are decoupled from the zero-shot algorithm \citep{touati2021learning, touati2022does}, mostly involving exploration policies trained with an intrinsic exploration reward~\citep{eysenbach2018diversity,burda2018exploration,lee2019efficient,liu2021aps,pathak2017curiosity,pathak2019self}.
Relevant to this work, \citet{chen2017ucb} proposed ensemble disagreement on the $Q$-value as an intrinsic reward for efficient exploration.
Alternatively, ensemble disagreement has been utilized in dynamics models for guiding exploration~\citep{pathak2017curiosity}.
In fact, subsequently, many works have successfully used this type of approach for exploration, connecting it with the notion of "epistemic uncertainty"~\citep{vlastelica2021risk, sukhija_optimistic_2023, sancaktar2022curious}.
While these methods yield successful exploration in some settings, a major disadvantage is that the exploration bonus doesn't depend on the rewards, so the exploration may focus on irrelevant aspects of the environment unrelated to the task \citep{chen2017ucb}. \looseness -1

A key question of this work is \textit{how should we best interact with the environment to learn all optimal policies in the unsupervised RL setting sample efficiently}? 
We aim to collect samples that are most informative about the occupancy measure of optimal policies encoded by a zero-shot RL algorithm, in other words, we want to minimize the uncertainty over the occupancy measures.
To this end, for modeling the occupancy measures we utilize the learned \algo{FB} factorization of occupancies~\citep{touati2021learning} which also has a representation space of optimal policies.
Inspired by \citet{lakshminarayanan2017simple}, we model the posterior predictive uncertainty over the $F$ representation by utilizing an ensemble of $F$ representations.
Consequently, the disagreement of the ensemble is a measure of uncertainty over $F$.
Because of the mechanics of the \algo{FB} representations, this naturally translates to the predictive uncertainty over the value function $Q^\piz(s,a)$ for a particular policy $\piz$ parametrized by reward embedding $z$, which is a more useful notion of uncertainty.
Motivated by insights from \textit{Bayesian experimental design}, we introduce an exploration algorithm that samples policies that are greedy \wrt to the mean of the $Q^\piz$-posterior, but have highest uncertainty.
This can be seen as a variant of \textit{uncertainty sampling}~\citep{lewis94}.
Our empirical evaluation indicates that utilizing this notion of uncertainty significantly improves the sample complexity of \algo{FB} with a simple exploration algorithm. 


In summary, in this work we provide an epistemic-uncertainty-guided method for efficiently learning forward-backward representations that (i) exhibits zero-shot generalization in unsupervised RL, (ii) leads to sample efficiency gains compared to other exploration alternatives and (iii) compares favorably  to current \algo{FB} methods when evaluated on several benchmarks. 


\section{Related Work}\label{sec:related-work}

\paragraph{Unsupervised Reinforcement Learning.}
Zero-shot (unsupervised) reinforcement learning frameworks can be traced back to the concept of a successor representation~\citep{dayan1993improving}, which relies on inferring the discounted occupancy measure of all policies.
A direct extension of this are successor features~\citep{barreto2017successor}, where a feature map is a assumed which linearizes the reward \wrt a representation $z$, the main caveat being that the map needs to be a-priori specified.
Consequently, many extensions exist to learn the feature map~\citep{hansen2019fast,laskin2022cic}.
Orthogonally, several works attempt to infer diverse skills in an online~\citep{eysenbach2018diversity} or offline fashion, mostly optimizing for a mutual-information objective.
In contrast to the former, forward-backward representations assume a factorization of the occupancy measure, where $z$ encodes an optimal value function for a specific reward. These can be traced back to \citet{blier2021}, and subsequent works have shown their effectiveness in deep RL benchmarks~\citep{touati2021learning,touati2022does,pirotta2024fast,tirinzoni2025zeroshot}, also dealing with the offline estimation problem of the \algo{FB}~\citep{jeen2023zero}.
In contrast to successor features, there has been no proper analysis of exploration for learning \algo{FB} representations more efficiently.
Our work aims to fill this gap. Other lines of work, such as unsupervised goal-conditioned RL focus on discovering a wide range of goals and learning corresponding goal-reaching policies, which lead to diverse learned behaviors \citep{mendonca2021discovering, pitis2020maximum, bagaria2019option}.

\paragraph{Exploration in Reinforcement Learning.}
\citet{lee2019efficient} attempt to solve the exploration problem by inferring the state marginal distribution of the policy and trying to match it to a user-defined target distribution.
\citet{osband2016deep} propose ensembles of $Q$ values for exploration by uniformly sampling a $Q$ function and subsequently following a policy associated with it for exploration.
Several works have extended the classic upper confidence bound (UCB) exploration strategy to deep RL via ensemble methods \citep{chen2017ucb,lee2021sunrise}, with \citet{lee2021sunrise} additionally proposing to account for the error in $Q$-targets by down-weighting based on ensemble  disagreement.
\citet{sukhija2024maxinforl} utilize an $\epsilon$-greedy policy with picking a Boltzmann policy with a mutual-information term for the dynamics.
\citet{metelli2019propagating} propagate uncertainty over $Q$-values by constructing a TD update by Wasserstein barycenters $V$; they propose several variants for inferring a policy (mean estimation, particle sampling).
Our work fits into the realm of ensemble-based exploration techniques, however in the context of zero-shot RL.

\paragraph{Deep Bayesian Inference.} The problem of exploration is closely related to active learning~\citep{chaloner1995, settles2009active}, also known as experimental design in the statistics literature. Active learning methods that yield strong theoretical generally query data points based on information-theoretic criteria \citep{krause08, settles2009active, hanneke2014}. These methods have recently generalized to deep learning. Since exact Bayesian inference is computationally intractable for neural networks, a variety of approximations have been developed \citep{mackay1992bayesian, neal2012bayesian}. \citet{gal2017deep, chen2017ucb} propose more computationally efficient methods than Bayesian neural networks, such as Monte Carlo dropout as an approximation of the posterior of the model parameters \citep{gal2017deep} or closer to our work, ensemble of neural networks \citep{osband2016deep, chen2017ucb, lakshminarayanan2017simple} for predictive uncertainty quantification. Several recent works further leverage such uncertainty estimates for active fine-tuning of vision or action models \citep{hubotter2024transductive, bagatella2024active}.

\textbf{Unsupervised Skill Discovery (USD).}
The goal of USD is to extract task-agnostic behaviors from intrinsic rewards. 
Most existing USD methods leverage mutual information (MI) as an intrinsic reward to learn diverse and distinguishable skills.
\citet{eysenbach_diversity_2018, laskin2022cic}  maximize a lower bound on MI between skills and states in a model-free way via a learned discriminator, while \citet{sharma_dynamics-aware_2020} learns the transition dynamics.  To enhance state space coverage exploration, ~\citep{strouse_learning_2022_optimistic} presents optimistic exploration through discriminator ensembles.
An alternate line of work replaces the MI objective with a Wasserstein dependency measure (WDM). 
METRA~\cite{park_metra_2024} and its variants~\cite{park2022lipschitzconstrainedunsupervisedskilldiscovery, park2023controllabilityawareunsupervisedskilldiscovery, rho_language_2024} maximize the directed distance in a learned latent space, resulting in state-covering skills.
Further efforts have been made in constructing offline task-regularized USD algorithms by leveraging Fenchel duality~\citep{vlastelicaoffline,kolevdual}.
Most of these methods, in contrast to zero-shot RL, require additional computation at inference time, for example via skill finetuning \citep{eysenbach_diversity_2018} or MPC planning \citep{sharma_dynamics-aware_2020}.

\section{Background}\label{sec:background}

In this paper we will utilize the standard notion of a reward-free Markov Decision Process which is defined by a tuple $\gM = (\gS,\rho_0, \gA, \gP, \gamma)$, with state space $\gS$, initial state distribution $\rho_0$, action space $\gA$, transition kernel $\gP$ and discount factor $\gamma$.

For the MDP $\gM$, a policy $\pi: \gS \rightarrow \gA$ induces the successor measure $M^\pi$ \citep{blier2021} for any initial state-action pair $(s_0, a_0)$:
\begin{equation}\label{eq:occupancy}
    M^\pi(s_0, a_0, X) \coloneqq \sum_{t \ge 0} \gamma^t P((s_{t+1},a_{t+1})\in X \, \vert  \, s_0, a_0, \pi) \quad \forall X \subset \gS \times \gA.
\end{equation}
Given $M^\pi$ and a reward function $r: \gS \times \gA \rightarrow \mathbb{R}$, we may write the value function of the policy $\pi$ for reward $r$ simply as $Q^\pi_r(s,a) = \sum_{s',a'}M^\pi(s, a, s',a') r(s',a')$.

Given a representation space $Z=\sR^d$ and a family of policies $(\pi_z)_{z \in \gZ}$ parameterized by $z$, the \algo{FB} representation looks for representations
$F: \gS \times \gA \times Z \rightarrow Z $ and $B: \gS \times \gA \rightarrow Z$, such that the successor measure $M^{\pi_z}$ in \eqref{eq:occupancy} factorizes as:   
\begin{equation}\label{eq:fb}
     M^{\pi_z}(s_0, a_0, s, a) \approx \langle F (s_0, a_0, z),  B(s,a) \rangle, \quad \pi_z(s) = \argmax_{a \in \gA} \langle F(s,a,z),  z \rangle,
\end{equation}
where $z$ is the latent representation of the policy $\pi_z$ of dimension $d$.
Assuming \eqref{eq:fb} holds, then for any reward function $r$, the policy $\pi_{z_r}$, where $z_r \defeq \sum_{s,a\in \gS \times \gA} r(s,a) B(s,a)$, is optimal for $r$ with optimal Q-function $Q^*_r(s,a) = \langle F(s,a,z_r), z_r \rangle$, \ie the policy is guaranteed to be optimal for any reward function~ \citep{touati2021learning}[Theorem 2].

In practice, we choose a parametric model $F_\theta$ and $B_\phi$ for the $F$ and $B$ representations, as approximations to the true successor measure 
factorization.
There are several off-the-shelf algorithms for learning $M^{\pi_z}$ \citep{blier2021, eysenbachc},
however the quality of the representation is tightly coupled with 1) the chosen factorization dimension $d$ and 2) the approximation error which can be result of model miss-specification or lack of data.
In this work we attempt to tackle the second issue, which can be handled by quantifying posterior uncertainty and utilizing it to guide exploration, similarly to previous works ~\citep{osband2013more,osband2016deep,chen2017ucb}.

\subsection{Bayesian Reinforcement Learning}

In the setting of Bayesian inference, ideally one would be able to formulate a prior distribution  over the parameters of the \algo{FB} representation $\Theta=(\theta,\phi)$ and subsequently, given evidence in form of data at the $i$-th iteration, compute the posterior distribution via Bayes' rule ${p(\Theta \vert \gD_i) = \frac{p(\Theta)p(\gD_i \vert \Theta)}{p(\gD_i)}}$.
This is intractable for high-dimensional $\Theta$, since it requires computing the marginal $p(\gD_i) = \int_\Theta p(\gD_i \vert \Theta) p(\Theta) d\Theta$.
Hence, many works have utilized various approximations to posterior distributions over neural networks~\citep{blundell2015weight,osband2016deep,chen2017ucb}.

Beyond proper quantification of uncertainty over $\Theta$, which is typically taken to be as variance or entropy of  $\Theta \sim p(\Theta \vert \gD_i)$ for continuous $\Theta$, the uncertainty of the predictions is of crucial interest in optimal data collection, which is a fundamental question in \emph{active learning} and \emph{optimal experimental design}.
In the field of \emph{bandits} and \emph{reinforcement learning}, this is also familiar under the term \emph{exploration}.
For RL in particular, one might want to compute a posterior over the unknown reward function $r$ and transition kernel $\gP$ of the MDP~\citep{osband2013more} or parameters of the $Q$-value function -- this is often approximated as an ensemble of neural networks in deep reinforcement learning~\citep{osband2016deep,chen2017ucb}.
Subsequently, this posterior is utilized in formulating an exploration strategy by a policy, a popular choice being a Upper-Confidence Bound (UCB) strategy by setting the policy to be $\pi(s) \defeq \argmax_{a \in \gA} \bar Q(s,a) + \alpha \sqrt{\Var[Q(s,a) \vert s,a]}$~\citep{chen2017ucb}, encouraging more uncertain actions.
This strategy stems from the well-known UCB algorithm in the bandit literature~\citep{auer2002finite,auer2002using}.
Further strategies exist that have been utilized in the literature, such as Thompson sampling where a particle is sampled from the posterior and subsequently exploited~\citep{osband2013more,thompson1933likelihood}.
\begin{remark}
    All of these methods focus on the exploitation-exploration tradeoff, which is ill-defined in the context of unsupervised reinforcement learning. This problem is fundamentally a \emph{pure exploration problem}.
\end{remark}

\section{Posterior uncertainty in forward-backward representations}\label{sec:epistemic_uncertainty}

In the unsupervised RL setting, accurately estimating the successor measure for all policies is of crucial interest. Given our prior distribution  over the parameters of the FB representation $\Theta=(\theta,\phi)$, we are tasked with updating the posterior distribution over the parameters as new evidence is collected.

Building on prior work that has successfully leveraged ensembles to approximate the posterior distribution over $Q^*$~\citep{osband2016deep,chen2017ucb}, we consider a similar approach for the FB representation.
Crucially, \citet{chen2017ucb} suggested decoupled $Q$ networks trained with standard 1-step TD error in order to approximate a posterior distribution given data $\gD$.
The FB representation entails a factorization of $M$ into $F$ and $B$, therefore naturally we might be tempted to construct a posterior over $F$ and $B$.
This however can cause issues, especially when utilizing ensemble methods, since the representation is non-unique (details can be found in \citet{blier2021}).
This is easy to see if we view the $F$ and $B$ functions as matrices: assuming a rotation matrix $R$, we have that $M = F^\top R R^{-1} B = \tilde F^\top \tilde B$, \ie $\tilde F$ and $\tilde B$ encode the same set of occupancy measures, however with the representation space rotated.
We alleviate this problem by fixing $B$ and modeling the posterior distribution  over $F$ alone.

Following \citet{chen2017ucb}, we adopt a naive posterior update over $F$:
for the $k$-th ensemble member in $[0, \dots, K-1]$, we minimize the empirical forward-backward loss over a batch of $b$ sampled transitions $(s_i, a_i, s_{i+1})_{i=0}^{b-1}$, independently sampled future states $(s_i')_{i=0}^{b-1}$, and reward embeddings $z_i$,
\def\blang{{\big\langle}}
\def\brang{{\big\rangle}}
\begin{align}\label{eq:posterior-update}
    \ell(\theta_k, \phi) =& \frac{1}{2b^2}\!\!\!\! \sum_{0 \leq i,j < b-1}\!\!\!\!\!\!\!\big (\blang F_{\theta_k}(s_i,a_i,z_i), B_\phi(s_j') \brang - \gamma \sum_{a \in \gA} \pi_{z_i}(a \qvert s_{i+1}) \blang F_{\theta_k^-}(s_{i+1}, a, z_i), B_{\phi-}(s_j')\brang\big )^2 \\&- 
    \frac{1}{b} \sum_{0 \leq i < b} \big \langle F_{\theta_k}(s_i,a_i,z_i), B_\phi(s_i)\big\rangle,
\end{align}
where $\theta_k^-$ and $\phi^-$ denote the target networks for $F_{\theta_k^-}$ and $B_{\phi^-}$ respectively.
In practice, an additional orthonormality regularization on $B$ is added as per \citet{touati2021learning} to normalize the covariance of $B$ (otherwise one could for example scale $F$ up and $B$ down since only $F^TB$ is fixed).

Equipped with a model to approximate the posterior distribution over forward representations, we are left with determining a strategy for collecting evidence to maximally reduce uncertainty of the posterior distribution, which is a challenging problem in deep learning. 
To design such an algorithm, we take inspiration from Bayesian experiment design \citep{macKay1992, chaloner1995}.
We shall adopt a well-known active learning heuristic -- \emph{uncertainty sampling}~\citep{lewis94}, which queries data points with the highest predictive uncertainty, but still provably minimizes posterior uncertainty under a homoscedastic, independent Gaussian noise model.

Aligned with previous work that showed that disagreement in ensemble methods can be effectively used for quantifying predictive uncertainty ~\citep{lakshminarayanan2017simple}, for a given query point $\boldsymbol{x}=(s,a,z)$, we model our distribution over $F$ as a uniformly weighted mixture of $\{F_k\}_{k=1}^{K}$ of Gaussian distributions \ie 
\begin{equation}\label{eq:posterior}
    p( F \qvert \boldsymbol{x}; \theta, \gD) \approx \frac{1}{K} \sum_k^K \gN( F ; \mu_{\theta_k}(\boldsymbol{x}), \Sigma_{\theta_k}(\boldsymbol{x})),
\end{equation}
where  $\boldsymbol{x} = (s,a,z)$ for ease of reading and $\mu_{\theta_k}, \Sigma_{\theta_k}$ are the mean and covariance predicted by ensemble member $k$ respectively.

In the limiting case of $\Sigma_{\theta_k} \rightarrow 0\; \forall k$ this posterior distribution becomes a mixture of Dirac delta functions, with the corresponding covariance being 
\begin{equation}\label{eq:variance}
    \mathrm{Cov}[F \qvert \boldsymbol{x};\theta, \gD] =  \frac{1}{K}\sum_k^K (F_k(\boldsymbol{x}) - \bar F(\boldsymbol{x})(F_k(\boldsymbol{x}) - \bar F(\boldsymbol{x}))^\top.
\end{equation}
where $F_k \defeq \mu_{\theta_k}$ and $\bar F \defeq \frac{1}{K}\sum_k \mu_{\theta_k}$.
While in previous work the variance of point estimates has been used in place of epistemic uncertainty exploration~\citep{lakshminarayanan2017simple}, here we have a matrix quantity, the covariance of the $F$-representations.
One viable option is to measure the volume of \cref{eq:variance}, by computing $\mathrm{Det}(\mathrm{Cov}[F \qvert s,a,z, \gD])$, the trace or maximum eigenvalue. We refer the reader to \cref{subsec:main-funcertainty} and \cref{app:f-uncertainty} for further analysis on this strategy, which we name \algo{FBEE$^F$}.
There is however, an argument against using \cref{eq:variance} to quantify uncertainty to guide data collection.
The primary object of interest for us is $Q^\piz$ for extracting greedy policies $\piz$ that are optimal \wrt some reward.
We may utilize the relationship $Q^{\pi_z}=\langle F(s,a,z), z \rangle$, to project our $F$-posterior to a $Q$-posterior, to arrive to the $Q$-predictive uncertainty for the query sample $(s,a,z)$.
\begin{equation}
    \Var[Q^{\pi_z}(s,a) \qvert \gD] = \frac{1}{K}\sum_{i=0}^K\langle F_k(s,a,z)-\bar F(s,a,z), z \rangle^2 .
\label{eq:q-variance}
\end{equation}
This corresponds to a Gaussian approximation to predictive posteriors on  $Q^\piz$.
In case of a Gaussian posterior, we have that its entropy is monotonic \wrt the variance, \ie for the case  of \cref{eq:q-variance} the predictive variance can be seen as a measure of information for the input query $\boldsymbol{x}$.
It is worth noting that because of the non-trivial dependence between $F^\piz$ and $z$, it is unclear how the predictive uncertainty of $F^\piz$ will affect the uncertainty on $Q^\piz$, which might be lower or higher after the projection with $z$.
This has the important consequence that minimizing the uncertainty on one versus the other may lead to significantly different algorithmic behaviors (see \cref{subsec:main-funcertainty} and \cref{app:f-uncertainty}).


\section{Epistemic exploration for FB representations}\label{sec:epistemic-exploration}

\begin{figure}[t!]
    \centering
    \includegraphics[width=0.65\linewidth]{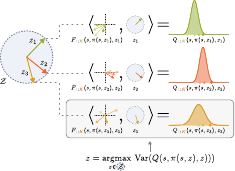}
    \caption{Epistemically guided \algo{FB} exploration (\method). During exploration we uniformly sample reward embeddings from a hypersphere (left), and take samples over our posterior distribution $F$ as represented by the $K$ ensemble members $F_{1:K}$ ($K=4$ in the figure) (middle-left). Then we project our $F$-posterior to a $Q$-posterior via $Q^{\pi_z}=\langle F(s,\piz(s),z), z \rangle$ (middle-right) and compute the Q-predictive uncertainty for all sampled $z$'s (left) via ensemble disagreement. We finally explore with the reward embedding $z^E$ that has maximum Q-predictive uncertainty.}
    \label{fig:descriptive-teaser}
\end{figure}

While we have a notion of posterior uncertainty phrased as the variance of the empirical predictive Q-posterior distribution in \cref{eq:q-variance}, it is still unclear how one can best  formulate an exploration policy for collecting data to improve the \algo{FB} representation.
To design such an algorithm, we take inspiration from \emph{Bayesian experimental design} \citep{chaloner1995,macKay1992}. 
A natural objective for active exploration is  maximizing mutual information between  $F$ and observed transition data $\gD_{i}$, which quantifies the reduction in entropy of $F$ conditioned on the observations.
In certain settings the predictive posterior variance is shown to be proportional to information gain~\citep{macKay1992}, hence it is a reasonable guide for exploration.

In general, we are seeking to define an exploration policy $\pi^E$ which is going to extend $\gD_{1:n-1}$ to $\gD_{1:n}$ such that the collected data $\gD_n$ provides the most amount of information about $F^\piz$ for all $\{\pi_z\}_{z \in \gZ}$. 
To this end, we take the approach of selecting a $\pi_z$ given $s,a$ that we are most uncertain about in terms of predictive variance, which may be seen as a variant of \emph{uncertainty sampling}, 
\begin{equation}\label{eq:exp-policy}
    \pi^E = \argmax_{\piz} \Var\big[\E_{a \sim \piz(s)}[Q^\piz(s,a)] \qvert s,a,z]\big] \quad \mathrm{s.t.}  \quad  z \in \gZ,
\end{equation}
where we make use of the posterior predictive variance in \cref{eq:q-variance}, which captures the uncertainty of the future return of $\piz$.
Although the exploration policy in $\cref{eq:exp-policy}$ is a greedy policy $\wrt$ $\langle \bar F^{\piz}(s,a), z \rangle$, we can still expect that executing $\piz$ reduces the uncertainty over $Q^\piz$.
Moreover, the uncertainty of different $Q$-posteriors depends on $z$ in a non-trivial way via $F^\piz$, hence a reduction in uncertainty in $Q^\piz$ is likely to reduce uncertainty across multiple $z \in \gZ$. 
This is loosely motivated by the "information never hurts" principle, which is a consequence of monotonicity of entropy $\gH[X \qvert Y] \le \gH[X]$ in light of new evidence $Y$. We provide pseudocode of our algorithm in \cref{algo:fb-epi} and a visual schematic in \cref{fig:descriptive-teaser}.

\begin{remark}
    While our definition of the policy in \cref{eq:exp-policy} is purely explorational, in the absence of a set of evaluation reward functions it is also reasonable, since there is no direct notion of "exploitation" in purely unsupervised RL.
\end{remark}

\begin{algorithm}
\caption{FB Uncertainty Sampling (\method)}\label{algo:fb-epi}
\begin{algorithmic}[1]
    \STATE \textbf{Input:} $K$-ensemble of $F_{\theta_k}$ and $F_{\theta_k^-}$, $B_\phi$ and $B_{\phi^-}$. 
    \WHILE{not converged}
        \STATE Pick $\pi^E$ according to \cref{eq:exp-policy}. 
        \STATE Collect data $\gD_n =\mathrm{Rollout}(\pi^E)$.
        \STATE Add data to buffer $\gD_{1:n} = \gD_{1:n-1} \cup \gD_n$.
        \STATE Fit $\{ F_{\theta_i}\}^K_{i=1}$, $B_\phi$ and policies $\piz$ with $\gD_{1:n}$.
    \ENDWHILE
\end{algorithmic}
\end{algorithm}



\section{Experiments}\label{sec:experiments}
Our experimental section is designed to provide an empirical answer to the following questions:
i) Does \method exhibit similar zero-shot generalization in online unsupervised RL compared to the original \algo{FB} method? ii) Does the epistemically guided exploration in \method lead to sample efficiency gains compared to other exploration alternatives? iii) What is the effect of exploring  over reward embeddings $z$'s compared to over actions? iv) How often should we update the chosen reward embedding $z^E$ during an exploration episode?  

\begin{figure}[t!]
\includegraphics[width=\linewidth]{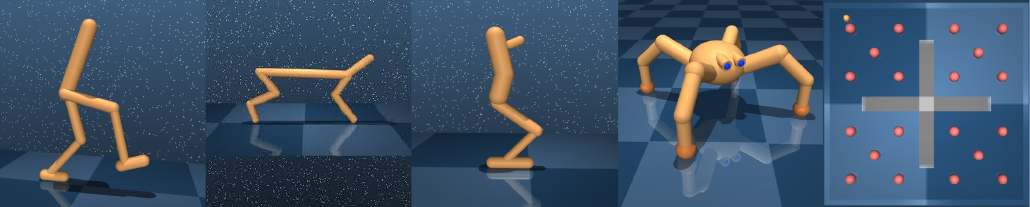}
\caption{Environments used in our experiments. (Left to right): \code{Walker, Cheetah, Hopper, Quadruped, Point-mass maze}. In the \code{Point-mass maze} domain, we show an example of initial state (yellow point), which always starts in the top-left room, and the 20 test goals (red circles).}
\vspace{-4mm}
\label{fig:dmc_envs}
\end{figure}

\fs{Environments}: We benchmark \method on 15 downstream tasks across 5 domains in the DeepMind Control Suite (DMC) \citep{tassa2018deepmind}, see \cref{fig:dmc_envs}). Details on the domains and tasks can be found in \cref{app:envs}.

\fs{Baselines}: We compare \method with several baselines for online unsupervised RL.
The first baseline is \algo{FB} \citep{touati2021learning}, the original FB algorithm that conducts uninformed exploration by uniformly sampling random reward embedding $z$'s.
We also compare against a naive \algo{random} policy that performs random exploration over the action space.
We additionally compare against \algo{FB-RND} \citep{touati2022does}, which decouples the exploration method from the learning of the \algo{FB} representation by leveraging a pure exploration method, namely \algo{RND} \citep{burda2018exploration}. We note that the exploration bonuses distilled by \algo{RND} remain independent of any estimate of FB representations. Notably, in this setting, we can leverage pre-collected exploration datasets from the Unsupervised Reinforcement Learning Benchmark \citep{laskin2021urlb}, and hence the FB representation is trained fully offline.
We also implement two variants of our algorithm:  \mbox{\method-\algo{policy}} explicitly learns an exploration policy $\pi_\theta: \mathcal{S} \to \mathcal{Z}$ by maximizing the objective in \cref{eq:exp-policy} through gradient descent, while \method-\algo{sampling} approximates the maximizer via zero-order optimization. Due to lack of space, we defer results of \method-\algo{policy} to the Appendix \cref{app:other-ablations}.
Finally, we implement an ablation of our method \algo{FBEE$^Q$-episode} to study the impact of how long to optimize for the most uncertain reward embedding $z^E$. With \algo{FBEE$^Q$-episode}, we only compute  $z^E$ (via \cref{eq:exp-policy}) at the beginning of each training episode, whereas the default implementation optimizes for it every 100 interaction steps (10 times more frequently). We implement this ablation for both \algo{FB} (\algo{FB-episode}) and our method \algo{FBEE$^Q$-episode}.


\paragraph{Results} We evaluate zero-shot performance of \method on 15 tasks across 5 domains in DMC every 100k exploration steps.
At evaluation time, given a task reward function $r(s,a)$, the agents acts with the reward representation  $z_R = \mathbb E_{(s, a) \sim \mathcal{D}} [r(s,a) B(s,a)]$ for 1000 environment steps. In practice, we compute the expectation by taking the average over relabeled samples from the current replay buffer. The reward function is bounded to [0,1], hence maximum return per task is of 1000. Zero-shot score curves averaged across tasks for every domain are shown in \cref{fig:all-performance-curves}. For zero-shot scores per each task, see \cref{fig:task-performance-curves-extra}.

As shown in \cref{fig:all-performance-curves}, \method asymptotically achieves similar or better performance than the original \algo{FB} method, hence answering our question i).
Most importantly, we observe that in all the environments \method exhibits significant sample efficiency gains compared to \algo{FB} across all domains, empirically showcasing that \method achieves the most important goal of our work, which is that of 
driving efficient exploration, hence answering our question ii). We notice that in easier tasks such as \code{cheetah} the performance gap between \algo{FB} and \method is reduced, showcasing that random exploration over reward embeddings is still a fairly good strategy. We leave a deeper  investigation of this finding to future work.
This naturally flows to answering our question iii) by which we empirically show that randomly exploring over reward embeddings leads to much sample efficiency than doing it at the action level. This can be observed by the low performance of the \algo{RANDOM} among all domains.

\newlength{\imw}
\setlength{\imw}{0.32\linewidth}
\begin{figure*}[!]
    \centering
    \begin{minipage}{\imw}
        \centering
        \includegraphics[width=\textwidth]{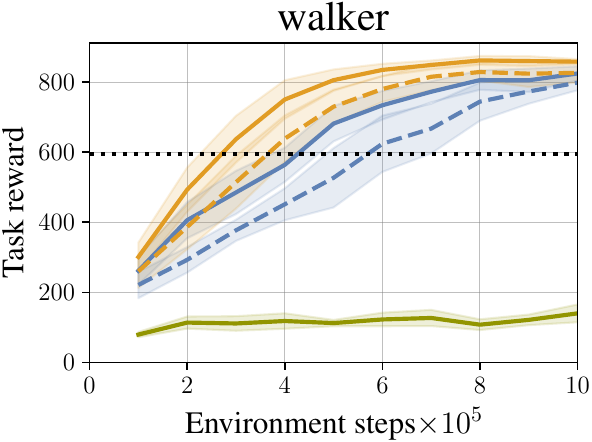}
    \end{minipage}
    \begin{minipage}{\imw}
        \centering
        \includegraphics[width=\textwidth]{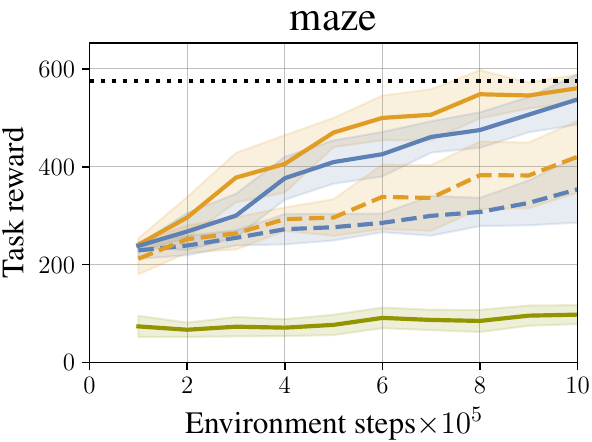}
    \end{minipage}
    \begin{minipage}{\imw}
        \centering
        \includegraphics[width=\textwidth]{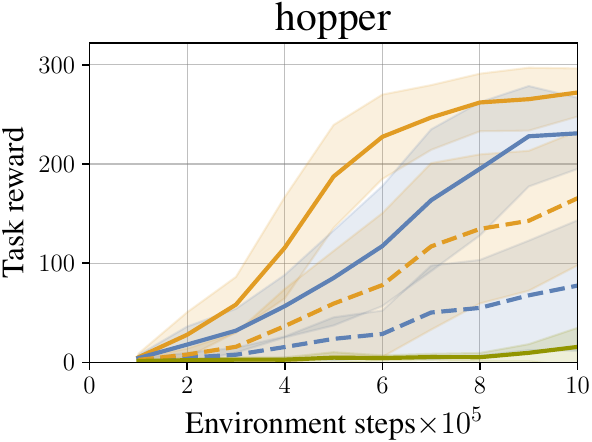}
    \end{minipage}

    \vspace{0.1cm}

    \begin{minipage}{\imw}
        \centering
        \includegraphics[width=\textwidth]{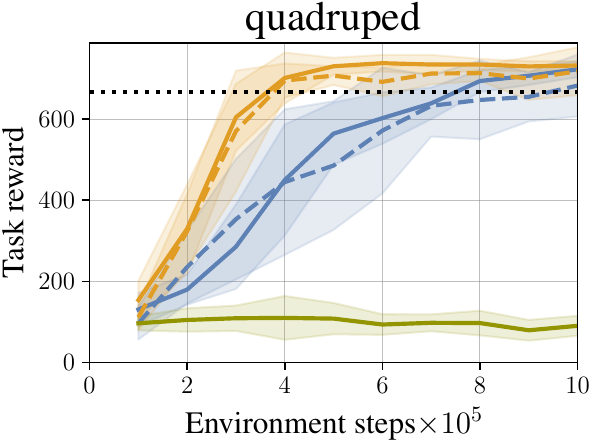}
    \end{minipage}
    \begin{minipage}{\imw}
        \centering
        \includegraphics[width=\textwidth]{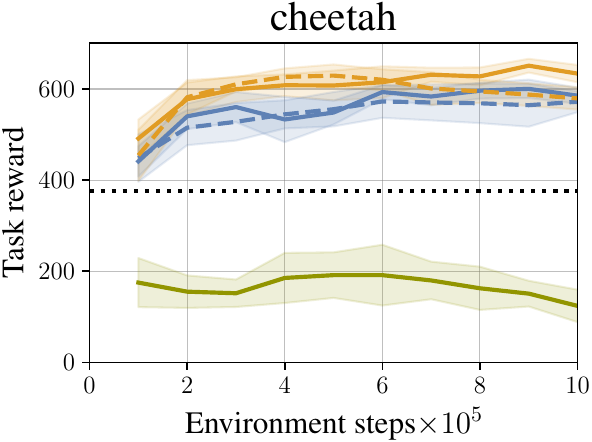}
    \end{minipage}
    \vspace{1em}
    

        \begin{tikzpicture}
            \draw[draw=none, fill=ourorange] (0,0.1) rectangle (0.5,0.2); 
            \node[right] at (0.6,0.15) {\small{\method}}; 
            
            \draw[draw=none, fill=ourblue] (2.5,0.1) rectangle (3.0,0.2); 
            \node[right] at (3.1,0.15) {\small{\algo{FB}}}; 
            
            \foreach \x in {4.0, 4.1, 4.2, 4.3, 4.4} { 
                \draw[draw=none, fill=ourorange] (\x,0.1) rectangle (\x+0.05,0.2); 
            }
            \node[right] at (4.55,0.15) {\small{\algo{FBEE$^Q$-episode}}}; %
            
            \foreach \x in {7.3, 7.4, 7.5, 7.6, 7.7} { 
            \draw[draw=none, fill=ourblue] (\x,0.1) rectangle (\x+0.05,0.2); 
        }
        \node[right] at (7.8,0.15) {\small{\algo{FB-episode}}}; %
        
        \draw[draw=none, fill=ourgreen] (10.1,0.1) rectangle (10.6,0.2); 
        \node[right] at (10.7,0.15) {\small{\algo{RANDOM}}}; 
       
        
        \end{tikzpicture}

    \caption{Zero-shot scores averaged over different downstream task as number of environment samples increases. Metrics are averaged over 30 evaluation episodes and 10 independent random seeds. Shaded area is 1-standard deviation. Topline is maximum score of \algo{FB-RND} (offline method with precollected data). Note: RND buffer for the \code{Hopper} task is not available in URLB benchmark \citep{laskin2021urlb}. }
    \vspace{-4mm}
    \label{fig:all-performance-curves}
\end{figure*}
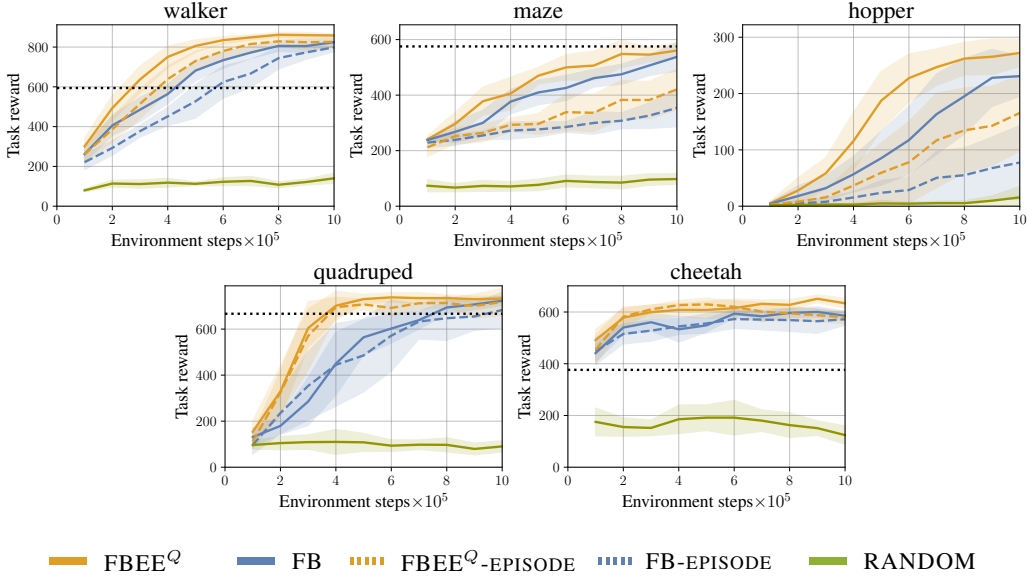

Finally, we are left with question iv), evaluating the impact on the $z^E$ frequency update during an exploration episode. We observe that for all methods the higher the frequency of the updates the better, although differences are only highly noticeable for the \code{hopper} and \code{maze} tasks.
For \method this could be due to several reasons. 
Our posterior update over $F$ differs from theoretically sound approaches (e.g., \citet{metelli2019propagating}, who suggested propagating uncertainty through a TD update involving Wasserstein barycenters) and can potentially incur myopic behavior. In practice, however, practical instantiations of similar algorithms \citep{metelli2019propagating} resort to the same approach as ours. Our hypothesis is that, as we update each of the ensemble members against its own target network, each member provides a temporally extended (and consistent) estimate of the value uncertainty via TD estimates, hence propagating uncertainty and alleviating myopic behavior. This was also observed by \citet{osband2016deep}.
\begin{wrapfigure}[12]{r}{0.4\textwidth}  
    \vspace{-4pt}
    \includegraphics[width=0.4\textwidth]{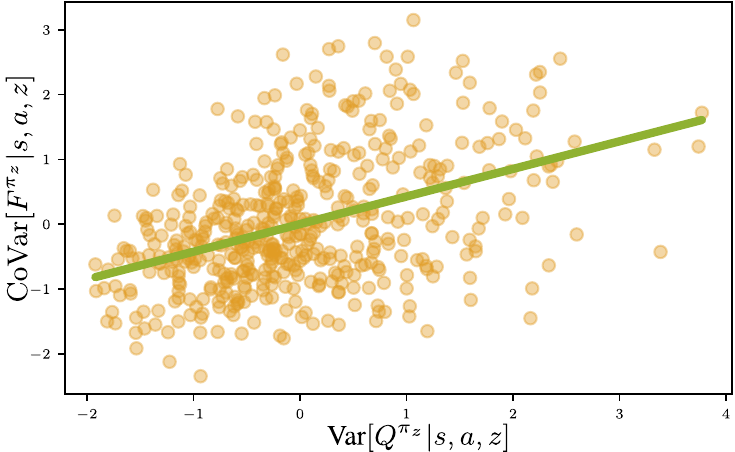}
    \caption{Regression scatter plot of the trace of $\mathrm{CoVar}[F^\piz \qvert s,a,z]$ and $\Var[Q^\piz \qvert s,a,z]$ for a \algo{FB} checkpoint in \code{Maze} experiment.}
    \label{fig:covar-var-plot}
\end{wrapfigure}
A second hypothesis would be that  our exploration strategy $\pi^E$ is not guaranteed to pick the $z^E$ that maximizes the cumulative posterior uncertainty over $Q$ \textit{over all $z$'s}, but instead is picking a $z$ that \textit{greedily} maximizes it. 
However, we empirically show that our method leads to significant sample efficiency gains compared to other exploration alternatives and we leave this analysis to future work.

\subsection{$F$-uncertainty versus $Q$-uncertainty.} \label{subsec:main-funcertainty} 
As we have argued in \cref{sec:epistemic_uncertainty}, $F^\piz$-uncertainty and $Q^\piz$-uncertainty may lead to different exploration behaviors. 
For purpose of demonstration, we analyze the average uncertainty across state-action pairs for the \code{Maze} experiment.
In \cref{fig:covar-var-plot} we observe how the  uncertainty of $F^\piz$ relates to the uncertainty of $Q^\piz$ for different $z$ samples in a particular \algo{FB} checkpoint from training.   Although there is a slight positive correlation between the trace of $\mathrm{CoVar}[F^\piz \qvert s,a,z]$ and $\Var[Q^\piz \qvert s,a,z]]$ in expectation, a low $R^2$ score of $0.18$ indicates that there is no strong correlation signal.
In fact, we observe instances where we have high $Q^\piz$ uncertainty and low $F^\piz$ uncertainty. 
To analyze the difference between the two exploration strategies, we implement a new variant of our algorithm, \algo{FBEE$^F$}, which uses predictive uncertainty in the $F$- representation to guide exploration (see \cref{app:f-uncertainty} for more details). We show that both strategies lead to similar performances, indicating that despite exploring the space differently, they are both seemingly valid exploration strategies.

\section{Conclusion}\label{sec:conclusion}

In this work we have proposed an epistemically-guided exploration framework for sample efficient learning of \algo{FB} representations.
We have done so by maintaining an ensemble approximation of the predictive posterior distribution over $Q^\piz$, and subsequently picking the least certain $\piz$ in terms of variance of $Q^\piz$, which can be seen as an instance of \emph{uncertainty sampling}.
This is a \emph{pure exploration} algorithm since the exploration-exploitation trade-off is non-existent in the zero-shot RL setting.
In experiments, this is an effective exploration strategy which outperforms other exploration algorithms on several environments of the DMC benchmark.

While this is a principled attempt at phrasing an exploration algorithm for zero-shot RL, many extensions are henceforth possible, such as extending this approach to further uncertainty-based algorithms such as UCB or Thompson sampling.
An efficient exploration algorithm necessarily needs to take into account how information is correlated across different $z \in \gZ$ in order to maximally reduce it with least amount of data.
Finally, a full Bayesian treatment of \algo{FB} representations is still an open question, especially with the assumption of a full posterior over $F$ and $B$, which is a difficult object because of the non-uniqueness of \algo{FB}.


\subsubsection*{Acknowledgments}
\label{sec:ack}
The authors thank Marco Bagatella for his advice during the project and his help in reviewing the manuscript and the anonymous reviewers for their valuable feedback.
The authors thank the Max Planck ETH Center for Learning Systems for supporting Núria Armengol and the ETH AI Center for supporting Marin Vlastelica.

Georg Martius is funded by the European Union (ERC, REAL-RL, 101045454). Views and opinions expressed are however those of the author(s) only and do not necessarily reflect those of the European Union or the European Research Council. Neither the European Union nor the granting authority can be held responsible for them. This work was also supported by the German Federal Ministry of Education and Research (BMBF): Tübingen AI Center, FKZ: 01IS18039A.


\newpage
\bibliography{main}
\bibliographystyle{rlj}

\beginSupplementaryMaterials

\input{appendix}
\end{document}

%% file: al_group_header.tex
\usepackage{times}
\usepackage[utf8]{inputenc} 
\usepackage[T1]{fontenc}    
\usepackage{url}            
\usepackage{booktabs}       
\usepackage{nicefrac}       
\usepackage{microtype}      
\usepackage{graphics,color}

\usepackage{algorithm}
\usepackage{algorithmic}
\usepackage{enumitem}

\usepackage{amsfonts}       
\usepackage{amsmath}       
\usepackage{amssymb}

\newcommand{\fs}[1]{{\bf #1}}

\usepackage{xspace}
\makeatletter
\DeclareRobustCommand\onedot{\futurelet\@let@token\@onedot}
\def\@onedot{\ifx\@let@token.\else.\null\fi\xspace}
\makeatother

\newcommand{\ie}{i.e\onedot}

\newcommand{\wrt}{w.r.t\onedot}
\newcommand{\defeq}{\vcentcolon=}

\usepackage{xr-hyper}
\makeatletter
\newcommand*{\addFileDependency}[1]{
  \typeout{(#1)}
  \@addtofilelist{#1}
  \IfFileExists{#1}{}{\typeout{No file #1.}}
}
\makeatother

\usepackage{xcolor}
\definecolor{ourblue}{rgb}{0.368,0.507,0.71}
\definecolor{ourorange}{rgb}{0.881,0.611,0.142}
\definecolor{ourgreen}{rgb}{0.56,0.692,0.195}
\definecolor{ourred}{rgb}{0.923,0.386,0.209}
\definecolor{ourviolet}{rgb}{0.528,0.471,0.701}
\definecolor{ourbrown}{rgb}{0.772,0.432,0.102}
\definecolor{ourlightblue}{rgb}{0.364,0.619,0.782}
\definecolor{ourdarkolive}{rgb}{0.572,0.586,0.}
\definecolor{ourdarkorange}{rgb}{0.71, 0.49, 0.1}
\definecolor{ourdarkblue}{rgb}{0.27, 0.4, 0.58}
\definecolor{ourdarkgreen}{rgb}{0.41, 0.51, 0.15}

\definecolor{ourcyan2}{rgb}{0.125,0.722,0.804}
\definecolor{ourred2}{rgb}{0.863,0.184,0.047}
\definecolor{ouryellow2}{cmyk}{0,0.16,1.0,0.07}
\definecolor{ourviolet2}{cmyk}{0.55,0.56,0,0.47}
\definecolor{ourorange2}{cmyk}{0,0.46,0.89,0.11}

\usepackage{multirow}

\usepackage{bbm}

\newcommand{\algo}[1]{\textsc{#1}}

\def\code#1{\texttt{#1}}

\newcounter{inlineequation}

%% file: math_commands.tex
\usepackage{amsmath,amsfonts,bm}

\def\1{\bm{1}}

\DeclareMathAlphabet{\mathsfit}{\encodingdefault}{\sfdefault}{m}{sl}
\SetMathAlphabet{\mathsfit}{bold}{\encodingdefault}{\sfdefault}{bx}{n}

\def\gA{{\mathcal{A}}}

\def\gD{{\mathcal{D}}}

\def\gH{{\mathcal{H}}}

\def\gM{{\mathcal{M}}}
\def\gN{{\mathcal{N}}}

\def\gP{{\mathcal{P}}}

\def\gS{{\mathcal{S}}}

\def\gZ{{\mathcal{Z}}}

\def\sR{{\mathbb{R}}}

\newcommand{\E}{\mathbb{E}}

\newcommand{\Var}{\mathrm{Var}}

\DeclareMathOperator*{\argmax}{arg\,max}

%% file: appendix.tex
\renewcommand{\thesection}{\Alph{section}} 
\setcounter{section}{0} 
\section{Environments} \label{app:envs}
All the environments are based on the \textit{DeepMind Control Suite}\citep{tassa2018deepmind} and some adapted by \citep{touati2022does}.
\begin{itemize}
\item \fs{Point-mass Maze}: a 2-dimensional continuous maze with four rooms. The states are 4-dimensional vectors encoding for positions and velocities of the point mass, and the actions are 2-dimensional vectors. Importantly, the initial position of the point-mass is  always sampled from a uniform distribution over the spatial domain of the top-left room only. At test, we evaluate performance of agents on 20 goal-reaching tasks (5 goals in each room described by their (x,y) coordinates (see \cref{fig:dmc_envs}).
This task is set as a goal-reaching task and hence we compute $z_R$ at evaluation time by: $z_R = B(s)$.
\item \fs{Cheetah}: A 17 state-dimensional running planar biped consisting of positions and velocities of robot joints. Actions are 6-dimensional. We evaluate on 4 tasks \code{walk, run, walk backward, run backward}. Rewards are linearly proportional to the achieved velocity up to the desired task velocity. 
\item \fs{Walker}: A 24 state-dimensional planar walker consisting of positions and velocities of robot joints. Actions are 6-dimensional. We evaluate on 4 tasks: \code{stand, run, flip}. In the \code{stand} task reward is a combination of terms encouraging an upright torso and some minimal
torso height. The \code{walk} and \code{run} task rewards include a component linearly proportional to the achieved velocity up to the desired task velocity. \code{flip} includes a component encouraging angular momentum.

\item \fs{Hopper}: A 15-dimensional planar one-legged hopper. Actions are 4 dimensional. We evaluate on 5 tasks: \code{stand, hop, flip}. In the \code{stand} the reward encourages a minimal torso height. In the \code{hop, hop backward} tasks the rewards have an additional term that is linearly proportional to the achieved velocity up to the desired task velocity. In the \code{flip, flip backward} includes a component encouraging angular momentum.


\item \fs{Quadruped} a four-leg spider navigating in 3D space. States and actions are 78 and 12 dimensional respectively. We evaluate on 4 tasks: \code{stand, walk, run jump}. \code{stand} reward encourages an upright torso, \code{walk} and \code{run} have an additional term that is linearly proportional to the achieved velocity up to the desired task velocity.  \code{jump} includes a term encouraging some minimal height of the center of mass.
\end{itemize}

\section{Prior information on rewards}
When dealing with high dimensionality environments, learning future probabilities for all states is very difficult and generally requires large $d$ to accommodate for all possible rewards.
In general, we are often interested in rewards that depend not on the full state but on a subset of it.
If this is known in advance, the representation B can be trained on that part of the state only, with same theoretical guarantees (Appendix, Theorem 4 \citep{touati2021learning}). Hence, when knowing that the reward will be only a function of a subset of the state and action spaces $G$, we can leverage an environment-dependent feature map $\varphi: S \times A \rightarrow G$, and learn $B(g)$ instead of $B(s,a)$, where $g=\varphi(s,a)$.
Importantly, rewards can be arbitrary functions of g. This was also suggested in \citep{touati2021learning}.
In what follows, we list the feature maps that were used for the different environments.
\begin{itemize}
\item \fs{Point-mass Maze}: $\phi(s,a) = [\code{$x,y$} ].$
\item \fs{Chetah}: $\phi(s,a)$ = [\code{$v_x,L_y$} ] where $v_x$ is the velocity along the x-axis in the robot frame and $L_x$ is the angular momentum about x-axis.
\item \fs{Walker}: $\phi(s,a)$ = [\code{$v_x, torso_z, torso_{z_w}$} ] where $v_x$ is the horizontal velocity of the center of mass, $torso_z$ is the height of the torso and $torso_{z_w}$ is the projection from the z-axis of the torso to the z-axis of the world frame.
\item \fs{Hopper}:  $\phi(s,a)$ = [\code{$v_x, torso_{z,foot}$} ] where  $v_x$ is the horizontal velocity of the center of mass and $torso_{z,foot}$ is the height of the torso with respect to the foot.
\item \fs{Humanoid}: $\phi(s,a)$ = [\code{$torso_z, v, torso_{z_w}$} ] where $torso_z$ is the height of the torso, v is the velocity of the center of mass in the local frame, and $torso_{z_w}$ is the projection from the z-axis of the torso to the z-axis of the world frame.
\item \fs{Quadruped} $\phi(s,a)$ = [\code{$v, torso_{z_w}$} ] where $v$ is the torso velocity vector in the local frame and $torso_{z_w}$ is the projection from the z-axis of the torso to the z-axis of the world frame.
\end{itemize}

\section{Hyperparameters}

In \cref{table:hyperparams} we summarize the hyperparameters used in our experiments. 
For a fair comparison, unless specified, we used the same parameters among all methods. Most of the parameters were adapted from \citep{touati2022does}.

\begin{table}[h]
    \centering
    \caption{Hyperparameters.}
    \label{table:hyperparams}

    \begin{tabular}{l l}
        \toprule
        \textbf{Hyperparameter} & \textbf{Value} \\
        \midrule
        Optimizer & Adam (default hyperparameters) \\
        Learning rate & $10^{-4}$ \\
        Batch size & $256$ \\
        Ratio gradient step/environment step & 0.5\\
        Z-dimension & $50$ ($100$ for maze) \\
        Discount factor $\gamma$ & $0.98$ ($0.99$ for maze) \\
        Mix ratio for $z$ sampling & $0.3$ \\
        Momentum coefficient for target networks update & $0.99$ \\
        Number of reward labels for task inference & $10^4$ \\
        Number of ensemble members & 5\\
        Frequency of z updates (training) & 0.01\\
        \bottomrule
    \end{tabular}
\end{table}

\section{Additional experiments}
\subsection{$F$-uncertainty versus $Q$-uncertainty exploration performance} \label{app:f-uncertainty}
As we have argued in \cref{sec:epistemic_uncertainty}, $F^\piz$-uncertainty and $Q^\piz$-uncertainty may lead to different exploration behaviors. In \cref{fig:covar-var-plot} we showed how, for a particular \algo{FB} checkpoint from training on the Maze experiment, there is not a strong correlation signal($R^2$ score of $0.18$) between the  uncertainty of $F^\piz$ to the uncertainty of $Q^\piz$.
In the following experiment, we compare the performance of \method with a new ablation (\algo{FBEE$^F$}), in which exploration is guided by the trace of the covariance of $F^{\pi_z}$. Specifically, we replace exploration as in \cref{eq:exp-policy} for:
\begin{equation}\label{eq:exp-policy-funcertainty}
    \pi^E = \argmax_{\piz} \text{tr}(\mathrm{CoVar}\big[\E_{a \sim \piz(s)}[F(s,a,z)]\big] \quad \mathrm{s.t.}  \quad  z \in \gZ.
\end{equation}
The results, shown in \cref{fig:all-performance-curves-funcertainty}, indicate that these two exploration strategies lead to similar performance, suggesting that using predictive uncertainty in the $F$-representation to guide data collection is a viable alternative. Per task performance scores are shown in \cref{fig:task-performance-curves-funcertainty}.

\setlength{\imw}{0.32\linewidth}
\begin{figure*}[!]
    \centering
    \begin{minipage}{\imw}
        \centering
        \includegraphics[width=\textwidth]{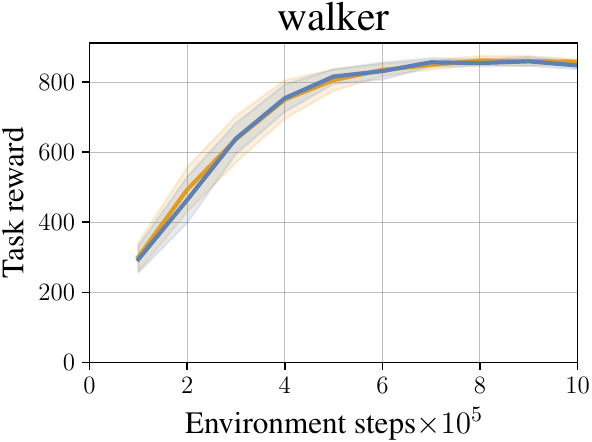}
    \end{minipage}
    \begin{minipage}{\imw}
        \centering
        \includegraphics[width=\textwidth]{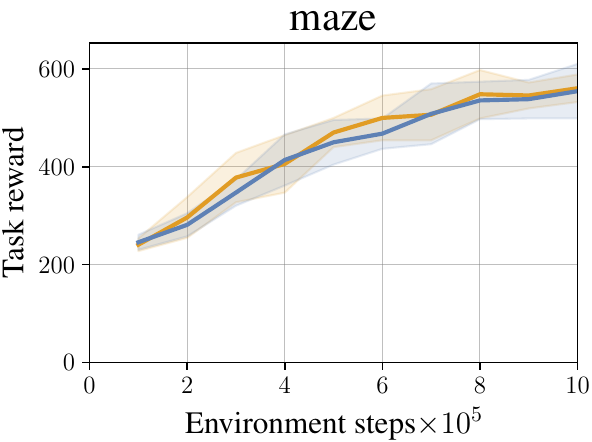}
    \end{minipage}
    \begin{minipage}{\imw}
        \centering
        \includegraphics[width=\textwidth]{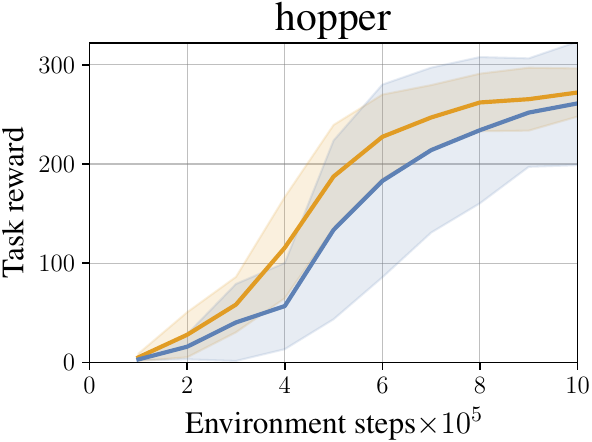}
    \end{minipage}

    \vspace{0.1cm}

    \begin{minipage}{\imw}
        \centering
        \includegraphics[width=\textwidth]{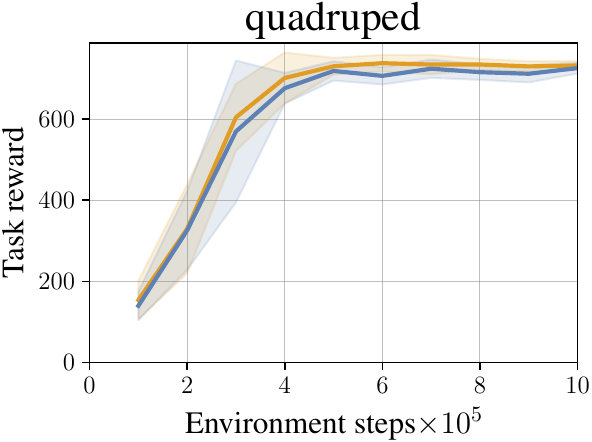}
    \end{minipage}
    \begin{minipage}{\imw}
        \centering
        \includegraphics[width=\textwidth]{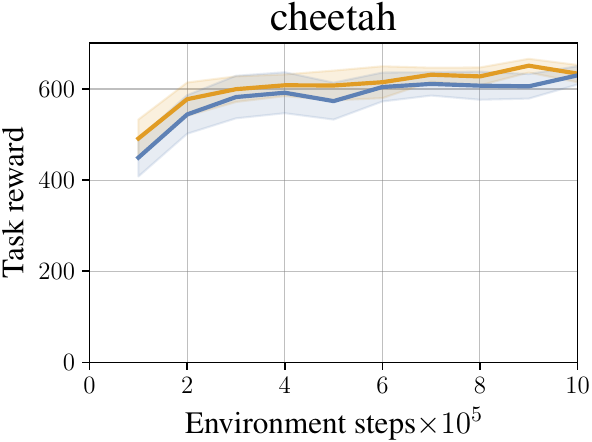}
    \end{minipage}
    \vspace{2em}
    

        \begin{tikzpicture}
        \draw[draw=none, fill=ourorange] (5.0,0.1) rectangle (5.45,0.2); 
        \node[right] at (5.5,0.15) {\small{\method}}; %
        
      \draw[draw=none, fill=ourblue] (7.0,0.1) rectangle (7.45,0.2); 
        \node[right] at (7.5,0.15) {\small{\algo{FBEE$^F$}}}; %
    
        \end{tikzpicture}

    \caption{Scores comparison when using  $F$-uncertainty versus $Q$-uncertainty exploration. Zero-shot scores averaged over different downstream task as number of environment samples increases. Metrics are averaged over 30 evaluation episodes and 10 independent random seeds. Shaded area is 1-standard deviation. }
    \label{fig:all-performance-curves-funcertainty}
\end{figure*}
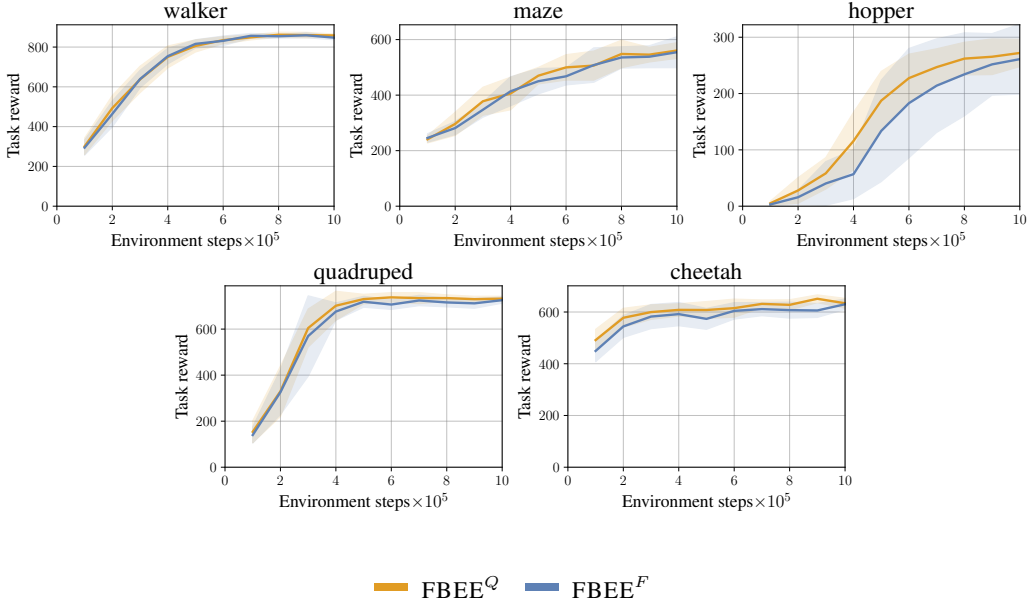

\subsection{Zero-shot scores per task}
\begin{figure*}[!]
    \centering
    \begin{minipage}{\textwidth}
        \centering
        \includegraphics[width=\linewidth]{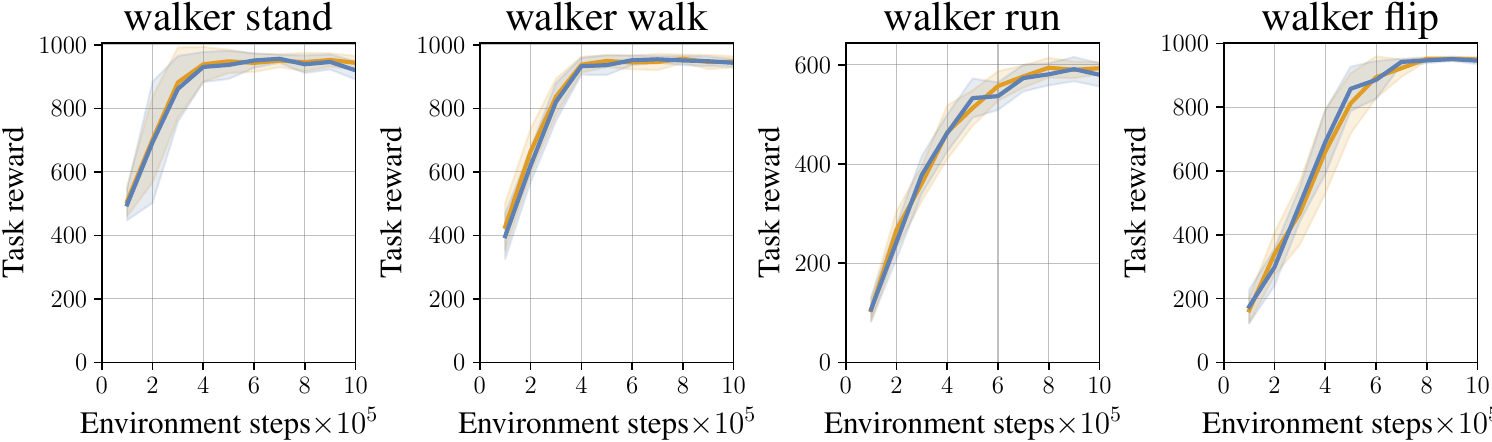}
    \end{minipage}%
    \vspace{0.01\textwidth} 

    \begin{minipage}{\textwidth}
        \centering
        \includegraphics[width=\linewidth]{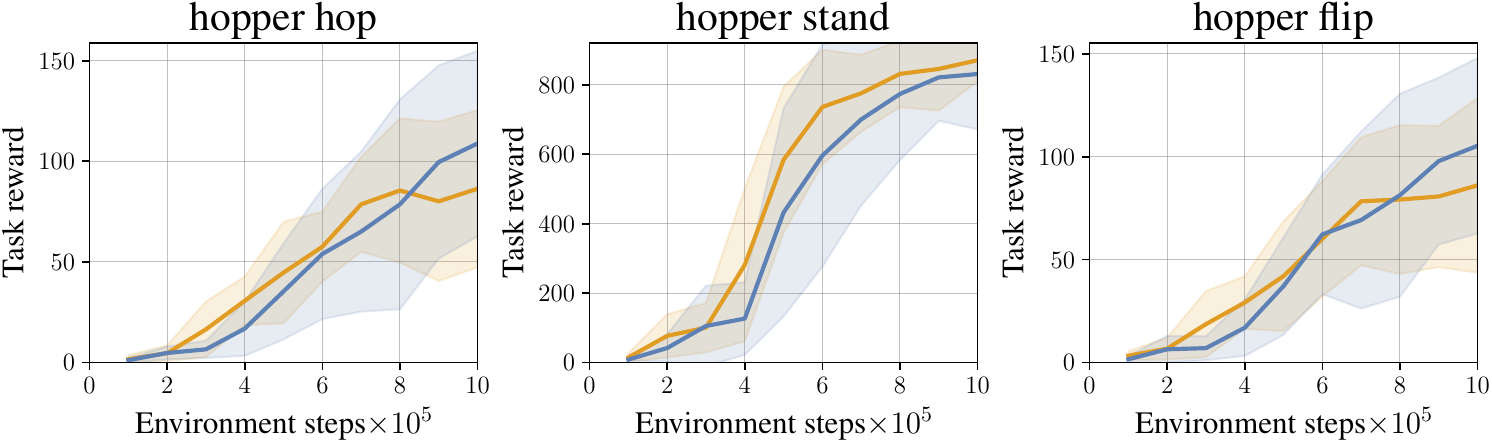}
    \end{minipage}
    \begin{minipage}{\textwidth}
        \centering
        \includegraphics[width=\linewidth]{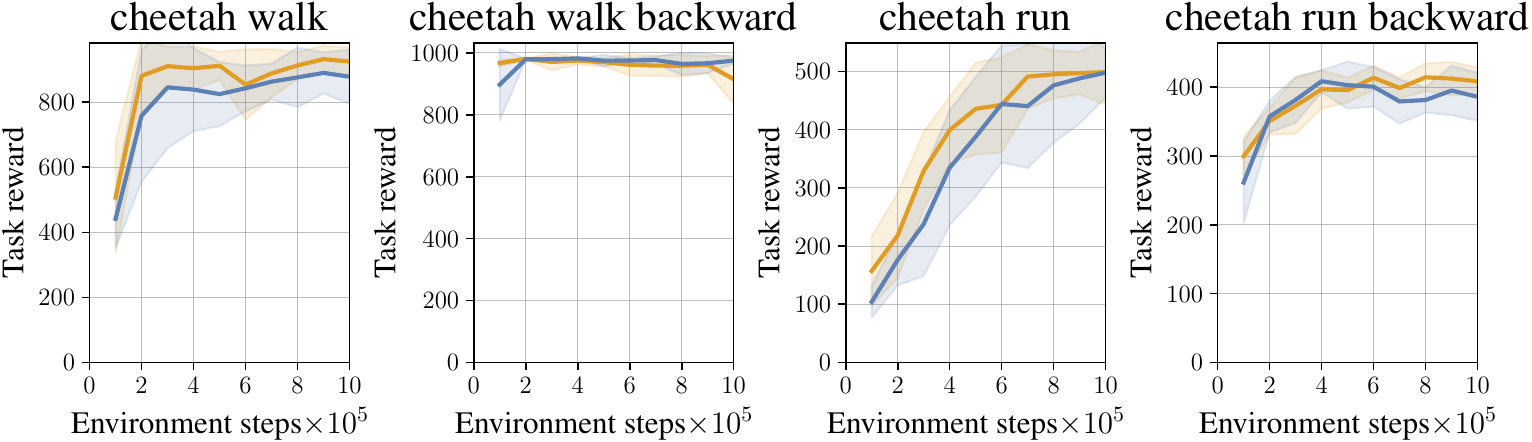}
    \end{minipage}%
    \vspace{0.01\textwidth} 

    \begin{minipage}{\textwidth}
        \centering
        \includegraphics[width=\linewidth]{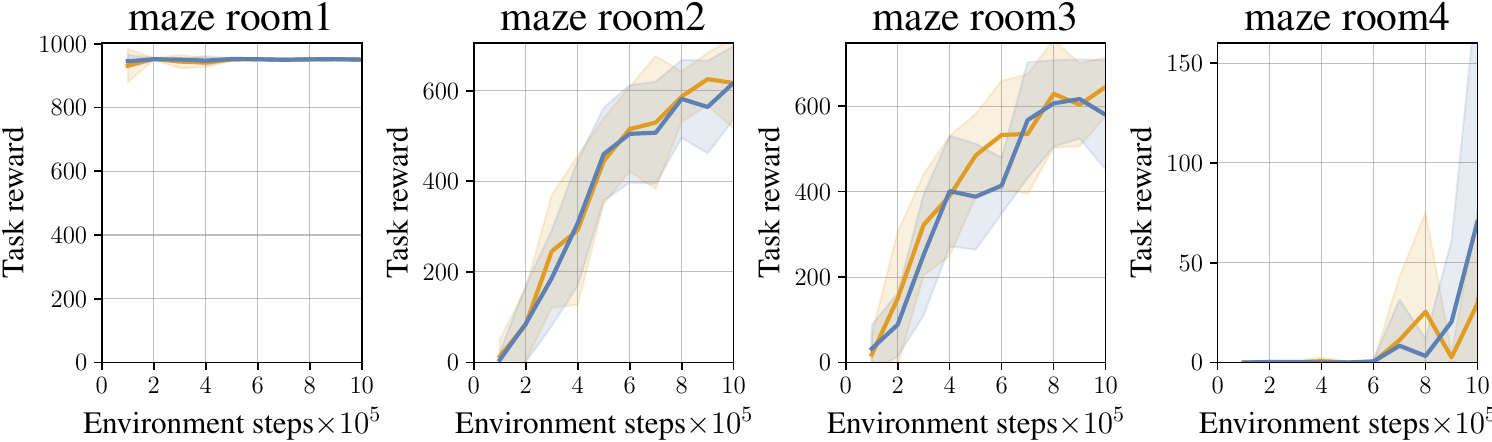}
    \end{minipage}
        \begin{minipage}{\textwidth}
        \centering
        \includegraphics[width=\linewidth]{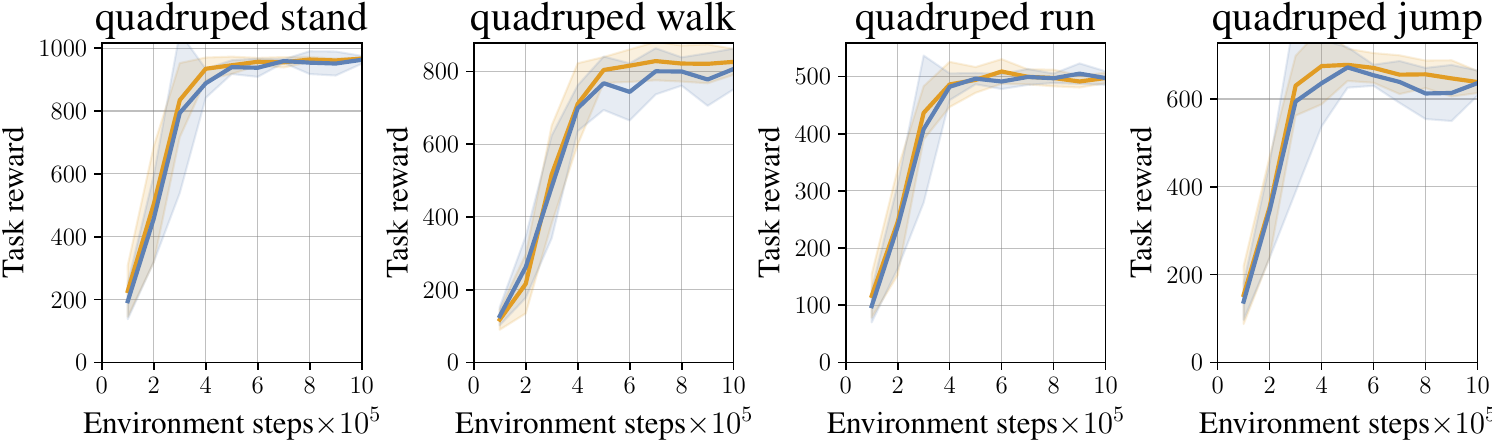}
    \end{minipage}
    \vspace{0.01\textwidth} 
    
        \begin{tikzpicture}
        \draw[draw=none, fill=ourorange] (5.0,0.1) rectangle (5.45,0.2); 
        \node[right] at (5.5,0.15) {\small{\method}}; %
        
      \draw[draw=none, fill=ourblue] (7.0,0.1) rectangle (7.45,0.2); 
        \node[right] at (7.5,0.15) {\small{\algo{FBEE$^F$}}}; %
    
        \end{tikzpicture}

    \caption{Zero-shot scores comparison when using  $F$-uncertainty versus $Q$-uncertainty exploration for different downstream task as number of environment samples increases. Metrics are averaged over 30 evaluation episodes and 10 independent random seeds. Shaded area is 1-standard deviation.}
    \label{fig:task-performance-curves-funcertainty}

\end{figure*}

We evaluate zero-shot performance of \method on 15 tasks across 5 domains in DMC every 100k exploration steps.
At evaluation time, given a task reward function $r(s,a)$, the agents acts with the reward representation  $z_R = \mathbb E [r(s,a) B(s,a)]$ for 1000 environment steps. The reward function is bounded to [0,1], hence maximum return per task is of 1000. In practice, we compute the expectation by taking the average over relabeled samples from the current replay buffer. Zero-shot scores across domains for all tasks is shown in \cref{fig:task-performance-curves-extra}. We also  show per task performance curves of the variant \algo{FBEE$^F$} in \cref{fig:task-performance-curves-funcertainty}.

\subsection{Other ablations}\label{app:other-ablations}
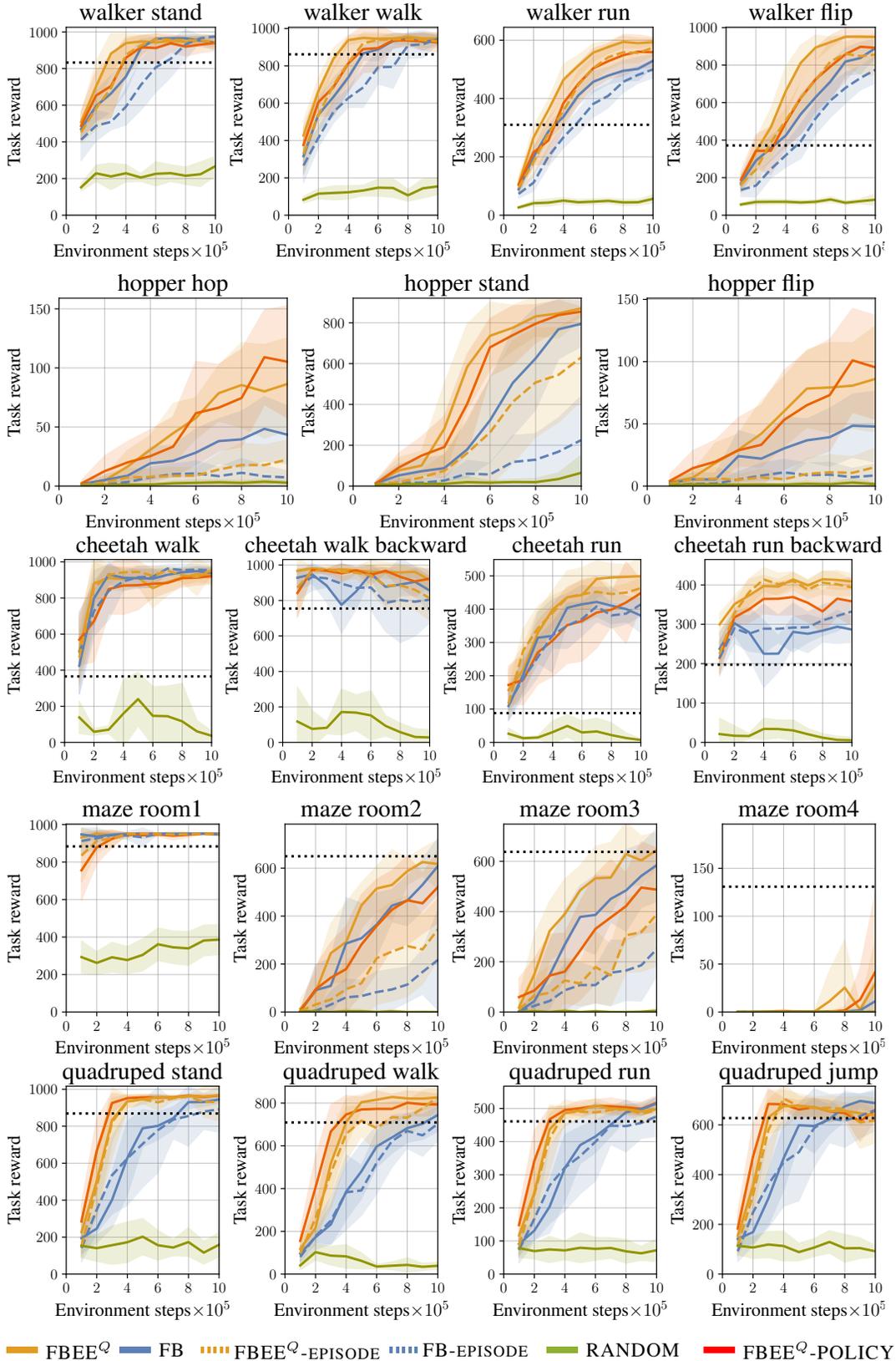
\begin{figure*}[!]
    \centering
    \begin{minipage}{\textwidth}
        \centering
        \includegraphics[width=\linewidth]{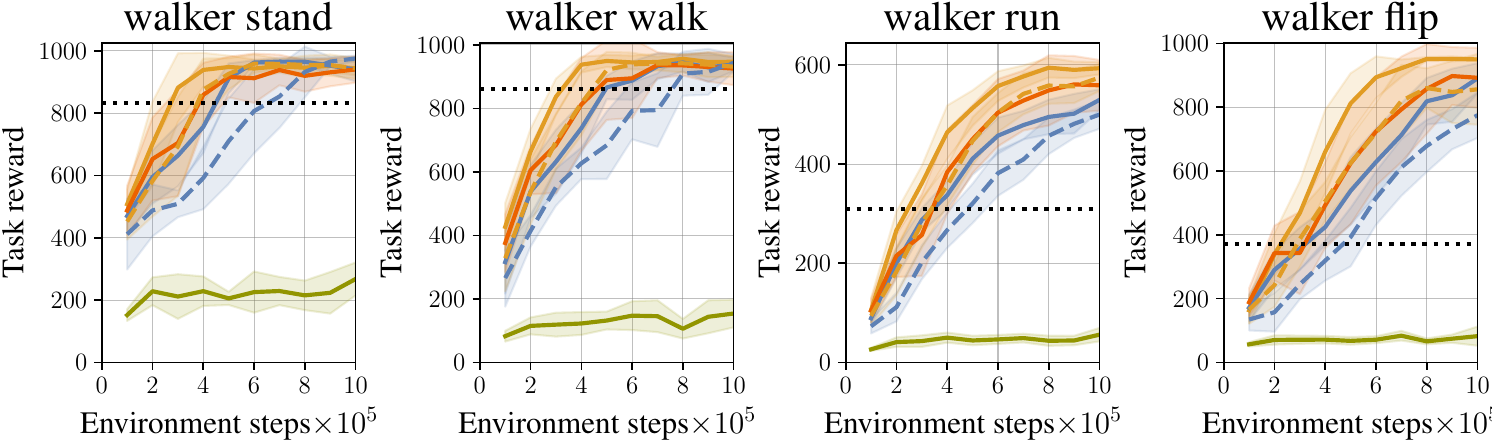}
    \end{minipage}%
    \vspace{0.01\textwidth} 

    \begin{minipage}{\textwidth}
        \centering
        \includegraphics[width=\linewidth]{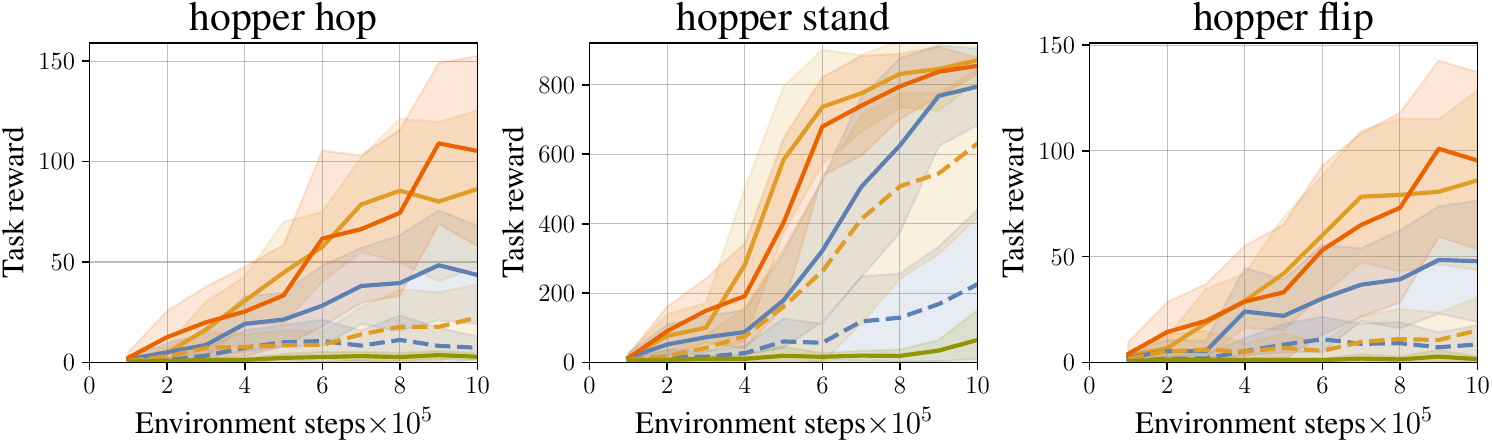}
    \end{minipage}
    \begin{minipage}{\textwidth}
        \centering
        \includegraphics[width=\linewidth]{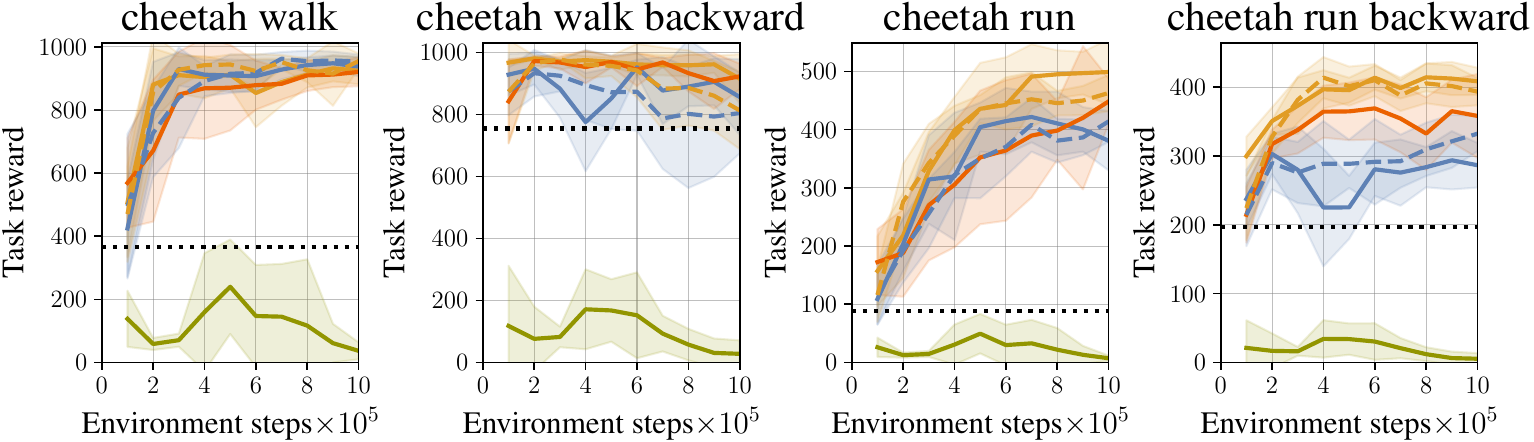}
    \end{minipage}%
    \vspace{0.01\textwidth} 

    \begin{minipage}{\textwidth}
        \centering
        \includegraphics[width=\linewidth]{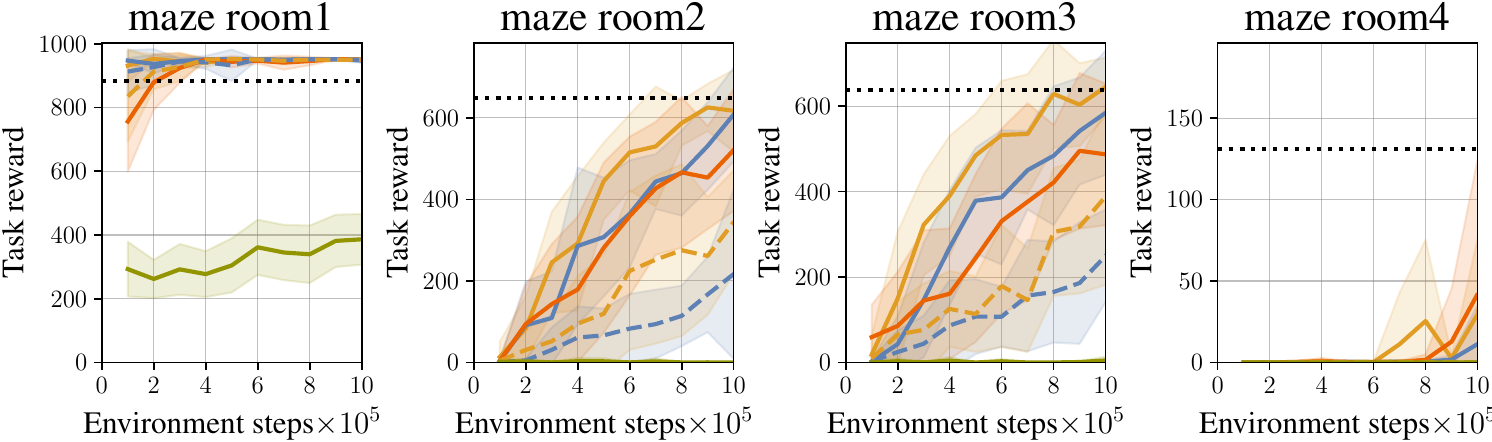}
    \end{minipage}
        \begin{minipage}{\textwidth}
        \centering
        \includegraphics[width=\linewidth]{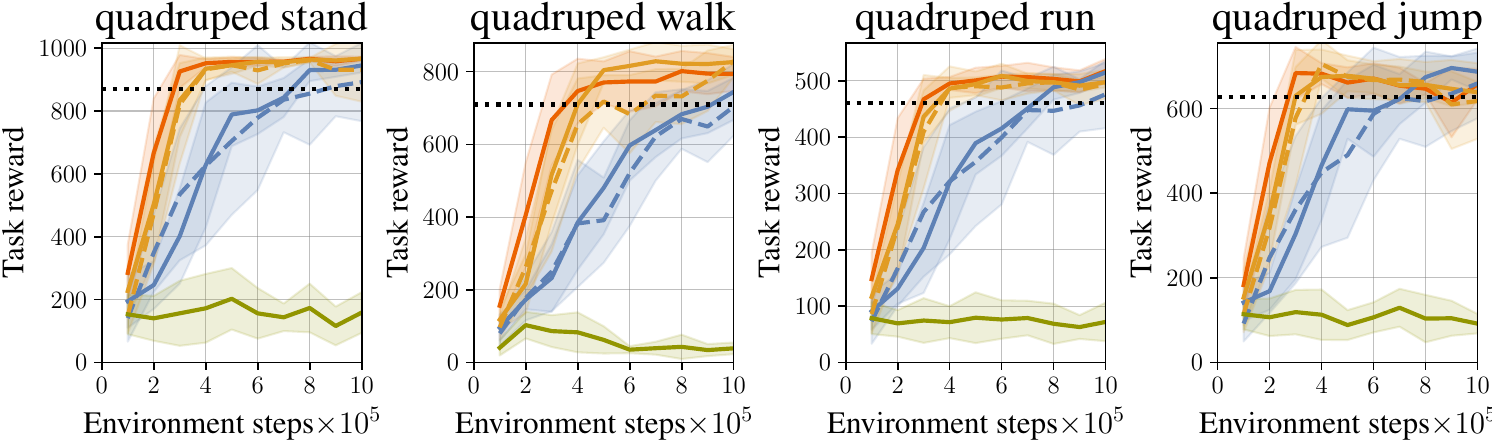}
    \end{minipage}
    \vspace{0.01\textwidth} 
    
     \begin{tikzpicture}
            \draw[draw=none, fill=ourorange] (0,0.1) rectangle (0.5,0.2); 
            \node[right] at (0.5,0.15) {\small{\method}}; 
            
            \draw[draw=none, fill=ourblue] (1.8,0.1) rectangle (2.3,0.2); 
            \node[right] at (2.3,0.15) {\small{\algo{FB}}}; 
            
            \foreach \x in {3.1, 3.2, 3.3, 3.4, 3.5} { 
                \draw[draw=none, fill=ourorange] (\x,0.1) rectangle (\x+0.05,0.2); 
            }
            \node[right] at (3.5,0.15) {\small{\algo{FBEE$^Q$-episode}}}; %
            
            \foreach \x in {6.1, 6.2, 6.3, 6.4, 6.5} { 
            \draw[draw=none, fill=ourblue] (\x,0.1) rectangle (\x+0.05,0.2); 
        }
        \node[right] at (6.5,0.15) {\small{\algo{FB-episode}}}; %
        
        \draw[draw=none, fill=ourgreen] (8.6,0.1) rectangle (9.1,0.2); 
        \node[right] at (9.1,0.15) {\small{\algo{RANDOM}}};

        \draw[draw=none, fill=ourdarkred] (11.1,0.1) rectangle (11.6,0.2); 
        \node[right] at (11.6,0.15) {\small{\algo{FBEE$^Q$-POLICY}}}; 
       

        \end{tikzpicture}
        
    \caption{Zero-shot scores for different downstream task as number of environment samples increases. Metrics are averaged over 30 evaluation episodes and 10 independent random seeds. Shaded area is 1-standard deviation. Topline is maximum score of \algo{FB-RND} (offline method with precollected data). Note: RND buffer for the \code{Hopper} task is not available in URLB benchmark \citep{laskin2021urlb}.}
    \label{fig:task-performance-curves-extra}

\end{figure*}

We additionally implement another ablation of our method, namely \method-\algo{policy} which explicitly learns an exploration policy $\pi_\theta: \mathcal{S} \to \mathcal{Z}$ by maximizing the objective in \cref{eq:exp-policy} through gradient descent. Results are shown in \cref{fig:task-performance-curves-extra}. In general we observe that it performs in par with \method-{\algo{sampling}}, and we attribute the mismatches in performance to not extensive hyperparameter finetuning.

%% file: main.bbl
\begin{thebibliography}{57}
\providecommand{\natexlab}[1]{#1}
\providecommand{\url}[1]{\texttt{#1}}
\expandafter\ifx\csname urlstyle\endcsname\relax
  \providecommand{\doi}[1]{DOI: #1}\else
  \providecommand{\doi}{DOI: \begingroup \urlstyle{rm}\Url}\fi

\bibitem[Auer(2002)]{auer2002using}
Peter Auer.
\newblock Using confidence bounds for exploitation-exploration trade-offs.
\newblock \emph{Journal of Machine Learning Research}, 3\penalty0 (Nov):\penalty0 397--422, 2002.

\bibitem[Auer et~al.(2002)Auer, Cesa-Bianchi, and Fischer]{auer2002finite}
Peter Auer, Nicolo Cesa-Bianchi, and Paul Fischer.
\newblock Finite-time analysis of the multiarmed bandit problem.
\newblock \emph{Machine learning}, 47:\penalty0 235--256, 2002.

\bibitem[Bagaria \& Konidaris(2019)Bagaria and Konidaris]{bagaria2019option}
Akhil Bagaria and George Konidaris.
\newblock Option discovery using deep skill chaining.
\newblock In \emph{International Conference on Learning Representations}, 2019.

\bibitem[Bagatella et~al.(2024)Bagatella, H{\"u}botter, Martius, and Krause]{bagatella2024active}
Marco Bagatella, Jonas H{\"u}botter, Georg Martius, and Andreas Krause.
\newblock Active fine-tuning of generalist policies.
\newblock \emph{arXiv preprint arXiv:2410.05026}, 2024.

\bibitem[Barreto et~al.(2017)Barreto, Dabney, Munos, Hunt, Schaul, van Hasselt, and Silver]{barreto2017successor}
Andr{\'e} Barreto, Will Dabney, R{\'e}mi Munos, Jonathan~J Hunt, Tom Schaul, Hado~P van Hasselt, and David Silver.
\newblock Successor features for transfer in reinforcement learning.
\newblock \emph{Advances in neural information processing systems}, 30, 2017.

\bibitem[Blier et~al.(2021)Blier, Tallec, and Ollivier]{blier2021}
L{\'{e}}onard Blier, Corentin Tallec, and Yann Ollivier.
\newblock Learning successor states and goal-dependent values: {A} mathematical viewpoint.
\newblock \emph{CoRR}, abs/2101.07123, 2021.
\newblock URL \url{https://arxiv.org/abs/2101.07123}.

\bibitem[Blundell et~al.(2015)Blundell, Cornebise, Kavukcuoglu, and Wierstra]{blundell2015weight}
Charles Blundell, Julien Cornebise, Koray Kavukcuoglu, and Daan Wierstra.
\newblock Weight uncertainty in neural network.
\newblock In \emph{International conference on machine learning}, pp.\  1613--1622. PMLR, 2015.

\bibitem[Burda et~al.(2018)Burda, Edwards, Storkey, and Klimov]{burda2018exploration}
Yuri Burda, Harrison Edwards, Amos Storkey, and Oleg Klimov.
\newblock Exploration by random network distillation.
\newblock \emph{arXiv preprint arXiv:1810.12894}, 2018.

\bibitem[Chaloner \& Verdinelli(1995)Chaloner and Verdinelli]{chaloner1995}
Kathryn Chaloner and Isabella Verdinelli.
\newblock {Bayesian Experimental Design: A Review}.
\newblock \emph{Statistical Science}, 10\penalty0 (3):\penalty0 273 -- 304, 1995.
\newblock \doi{10.1214/ss/1177009939}.
\newblock URL \url{https://doi.org/10.1214/ss/1177009939}.

\bibitem[Chen et~al.(2017)Chen, Sidor, Abbeel, and Schulman]{chen2017ucb}
Richard~Y Chen, Szymon Sidor, Pieter Abbeel, and John Schulman.
\newblock Ucb exploration via q-ensembles.
\newblock \emph{arXiv preprint arXiv:1706.01502}, 2017.

\bibitem[Dayan(1993)]{dayan1993improving}
Peter Dayan.
\newblock Improving generalization for temporal difference learning: The successor representation.
\newblock \emph{Neural computation}, 5\penalty0 (4):\penalty0 613--624, 1993.

\bibitem[Eysenbach et~al.(2018{\natexlab{a}})Eysenbach, Gupta, Ibarz, and Levine]{eysenbach2018diversity}
Benjamin Eysenbach, Abhishek Gupta, Julian Ibarz, and Sergey Levine.
\newblock Diversity is all you need: Learning skills without a reward function.
\newblock \emph{arXiv preprint arXiv:1802.06070}, 2018{\natexlab{a}}.

\bibitem[Eysenbach et~al.(2018{\natexlab{b}})Eysenbach, Gupta, Ibarz, and Levine]{eysenbach_diversity_2018}
Benjamin Eysenbach, Abhishek Gupta, Julian Ibarz, and Sergey Levine.
\newblock Diversity is {All} {You} {Need}: {Learning} {Skills} without a {Reward} {Function}, October 2018{\natexlab{b}}.
\newblock URL \url{http://arxiv.org/abs/1802.06070}.
\newblock arXiv:1802.06070.

\bibitem[Eysenbach et~al.(2021)Eysenbach, Salakhutdinov, and Levine]{eysenbachc}
Benjamin Eysenbach, Ruslan Salakhutdinov, and Sergey Levine.
\newblock C-learning: Learning to achieve goals via recursive classification.
\newblock In \emph{International Conference on Learning Representations}, 2021.

\bibitem[Gal et~al.(2017)Gal, Islam, and Ghahramani]{gal2017deep}
Yarin Gal, Riashat Islam, and Zoubin Ghahramani.
\newblock Deep bayesian active learning with image data.
\newblock In \emph{International conference on machine learning}, pp.\  1183--1192. PMLR, 2017.

\bibitem[Hanneke et~al.(2014)]{hanneke2014}
Steve Hanneke et~al.
\newblock Theory of disagreement-based active learning.
\newblock \emph{Foundations and Trends{\textregistered} in Machine Learning}, 7\penalty0 (2-3):\penalty0 131--309, 2014.

\bibitem[Hansen et~al.(2019)Hansen, Dabney, Barreto, Van~de Wiele, Warde-Farley, and Mnih]{hansen2019fast}
Steven Hansen, Will Dabney, Andre Barreto, Tom Van~de Wiele, David Warde-Farley, and Volodymyr Mnih.
\newblock Fast task inference with variational intrinsic successor features.
\newblock \emph{arXiv preprint arXiv:1906.05030}, 2019.

\bibitem[H{\"u}botter et~al.(2024)H{\"u}botter, Sukhija, Treven, As, and Krause]{hubotter2024transductive}
Jonas H{\"u}botter, Bhavya Sukhija, Lenart Treven, Yarden As, and Andreas Krause.
\newblock Transductive active learning: Theory and applications.
\newblock In \emph{The Thirty-eighth Annual Conference on Neural Information Processing Systems}, 2024.

\bibitem[Jeen et~al.(2023)Jeen, Bewley, and Cullen]{jeen2023zero}
Scott Jeen, Tom Bewley, and Jonathan~M Cullen.
\newblock Zero-shot reinforcement learning from low quality data.
\newblock \emph{arXiv preprint arXiv:2309.15178}, 2023.

\bibitem[Kolev et~al.(2025)Kolev, Vlastelica, and Martius]{kolevdual}
Pavel Kolev, Marin Vlastelica, and Georg Martius.
\newblock Dual-force: Enhanced offline diversity maximization under imitation constraints.
\newblock In \emph{Seventeenth European Workshop on Reinforcement Learning}, 2025.

\bibitem[Krause et~al.(2008)Krause, Singh, and Guestrin]{krause08}
Andreas Krause, Ajit Singh, and Carlos Guestrin.
\newblock Near-optimal sensor placements in gaussian processes: Theory, efficient algorithms and empirical studies.
\newblock \emph{Journal of Machine Learning Research}, 9\penalty0 (8):\penalty0 235--284, 2008.
\newblock URL \url{http://jmlr.org/papers/v9/krause08a.html}.

\bibitem[Lakshminarayanan et~al.(2017)Lakshminarayanan, Pritzel, and Blundell]{lakshminarayanan2017simple}
Balaji Lakshminarayanan, Alexander Pritzel, and Charles Blundell.
\newblock Simple and scalable predictive uncertainty estimation using deep ensembles.
\newblock \emph{Advances in neural information processing systems}, 30, 2017.

\bibitem[Laskin et~al.(2021)Laskin, Yarats, Liu, Lee, Zhan, Lu, Cang, Pinto, and Abbeel]{laskin2021urlb}
Michael Laskin, Denis Yarats, Hao Liu, Kimin Lee, Albert Zhan, Kevin Lu, Catherine Cang, Lerrel Pinto, and Pieter Abbeel.
\newblock Urlb: Unsupervised reinforcement learning benchmark.
\newblock \emph{arXiv preprint arXiv:2110.15191}, 2021.

\bibitem[Laskin et~al.(2022)Laskin, Liu, Peng, Yarats, Rajeswaran, and Abbeel]{laskin2022cic}
Michael Laskin, Hao Liu, Xue~Bin Peng, Denis Yarats, Aravind Rajeswaran, and Pieter Abbeel.
\newblock Cic: Contrastive intrinsic control for unsupervised skill discovery.
\newblock \emph{arXiv preprint arXiv:2202.00161}, 2022.

\bibitem[Lee et~al.(2021)Lee, Laskin, Srinivas, and Abbeel]{lee2021sunrise}
Kimin Lee, Michael Laskin, Aravind Srinivas, and Pieter Abbeel.
\newblock Sunrise: A simple unified framework for ensemble learning in deep reinforcement learning.
\newblock In \emph{International Conference on Machine Learning}, pp.\  6131--6141. PMLR, 2021.

\bibitem[Lee et~al.(2019)Lee, Eysenbach, Parisotto, Xing, Levine, and Salakhutdinov]{lee2019efficient}
Lisa Lee, Benjamin Eysenbach, Emilio Parisotto, Eric Xing, Sergey Levine, and Ruslan Salakhutdinov.
\newblock Efficient exploration via state marginal matching.
\newblock \emph{arXiv preprint arXiv:1906.05274}, 2019.

\bibitem[Lewis \& Gale(1994)Lewis and Gale]{lewis94}
David~D. Lewis and William~A. Gale.
\newblock A sequential algorithm for training text classifiers.
\newblock In Bruce~W. Croft and C.~J. van Rijsbergen (eds.), \emph{SIGIR '94}, pp.\  3--12, London, 1994. Springer London.
\newblock ISBN 978-1-4471-2099-5.

\bibitem[Liu \& Abbeel(2021)Liu and Abbeel]{liu2021aps}
Hao Liu and Pieter Abbeel.
\newblock Aps: Active pretraining with successor features.
\newblock In \emph{International Conference on Machine Learning}, pp.\  6736--6747. PMLR, 2021.

\bibitem[MacKay(1992)]{macKay1992}
David J.~C. MacKay.
\newblock Information-based objective functions for active data selection.
\newblock \emph{Neural Computation}, 4\penalty0 (4):\penalty0 590--604, 07 1992.
\newblock ISSN 0899-7667.
\newblock \doi{10.1162/neco.1992.4.4.590}.
\newblock URL \url{https://doi.org/10.1162/neco.1992.4.4.590}.

\bibitem[Mackay(1992)]{mackay1992bayesian}
David John~Cameron Mackay.
\newblock \emph{Bayesian methods for adaptive models}.
\newblock California Institute of Technology, 1992.

\bibitem[Mendonca et~al.(2021)Mendonca, Rybkin, Daniilidis, Hafner, and Pathak]{mendonca2021discovering}
Russell Mendonca, Oleh Rybkin, Kostas Daniilidis, Danijar Hafner, and Deepak Pathak.
\newblock Discovering and achieving goals via world models.
\newblock \emph{Advances in Neural Information Processing Systems}, 34:\penalty0 24379--24391, 2021.

\bibitem[Metelli et~al.(2019)Metelli, Likmeta, and Restelli]{metelli2019propagating}
Alberto~Maria Metelli, Amarildo Likmeta, and Marcello Restelli.
\newblock Propagating uncertainty in reinforcement learning via wasserstein barycenters.
\newblock In H.~Wallach, H.~Larochelle, A.~Beygelzimer, F.~d\textquotesingle Alch\'{e}-Buc, E.~Fox, and R.~Garnett (eds.), \emph{Advances in Neural Information Processing Systems}, volume~32. Curran Associates, Inc., 2019.
\newblock URL \url{https://proceedings.neurips.cc/paper_files/paper/2019/file/f83630579d055dc5843ae693e7cdafe0-Paper.pdf}.

\bibitem[Neal(2012)]{neal2012bayesian}
Radford~M Neal.
\newblock \emph{Bayesian learning for neural networks}, volume 118.
\newblock Springer Science \& Business Media, 2012.

\bibitem[Osband et~al.(2013)Osband, Russo, and Van~Roy]{osband2013more}
Ian Osband, Daniel Russo, and Benjamin Van~Roy.
\newblock (more) efficient reinforcement learning via posterior sampling.
\newblock \emph{Advances in Neural Information Processing Systems}, 26, 2013.

\bibitem[Osband et~al.(2016)Osband, Blundell, Pritzel, and Van~Roy]{osband2016deep}
Ian Osband, Charles Blundell, Alexander Pritzel, and Benjamin Van~Roy.
\newblock Deep exploration via bootstrapped dqn.
\newblock \emph{Advances in neural information processing systems}, 29, 2016.

\bibitem[Park et~al.(2022)Park, Choi, Kim, Lee, and Kim]{park2022lipschitzconstrainedunsupervisedskilldiscovery}
Seohong Park, Jongwook Choi, Jaekyeom Kim, Honglak Lee, and Gunhee Kim.
\newblock Lipschitz-constrained unsupervised skill discovery, 2022.
\newblock URL \url{https://arxiv.org/abs/2202.00914}.

\bibitem[Park et~al.(2023)Park, Lee, Lee, and Abbeel]{park2023controllabilityawareunsupervisedskilldiscovery}
Seohong Park, Kimin Lee, Youngwoon Lee, and Pieter Abbeel.
\newblock Controllability-aware unsupervised skill discovery, 2023.
\newblock URL \url{https://arxiv.org/abs/2302.05103}.

\bibitem[Park et~al.(2024)Park, Rybkin, and Levine]{park_metra_2024}
Seohong Park, Oleh Rybkin, and Sergey Levine.
\newblock {METRA}: {Scalable} {Unsupervised} {RL} with {Metric}-{Aware} {Abstraction}, March 2024.
\newblock URL \url{http://arxiv.org/abs/2310.08887}.
\newblock arXiv:2310.08887.

\bibitem[Pathak et~al.(2017)Pathak, Agrawal, Efros, and Darrell]{pathak2017curiosity}
Deepak Pathak, Pulkit Agrawal, Alexei~A Efros, and Trevor Darrell.
\newblock Curiosity-driven exploration by self-supervised prediction.
\newblock In \emph{International Conference on Machine Learning}, pp.\  2778--2787. PMLR, 2017.

\bibitem[Pathak et~al.(2019)Pathak, Gandhi, and Gupta]{pathak2019self}
Deepak Pathak, Dhiraj Gandhi, and Abhinav Gupta.
\newblock Self-supervised exploration via disagreement.
\newblock In \emph{International Conference on Machine Learning}, pp.\  5062--5071. PMLR, 2019.

\bibitem[Pirotta et~al.(2024)Pirotta, Tirinzoni, Touati, Lazaric, and Ollivier]{pirotta2024fast}
Matteo Pirotta, Andrea Tirinzoni, Ahmed Touati, Alessandro Lazaric, and Yann Ollivier.
\newblock Fast imitation via behavior foundation models.
\newblock In \emph{The Twelfth International Conference on Learning Representations}, 2024.

\bibitem[Pitis et~al.(2020)Pitis, Chan, Zhao, Stadie, and Ba]{pitis2020maximum}
Silviu Pitis, Harris Chan, Stephen Zhao, Bradly Stadie, and Jimmy Ba.
\newblock Maximum entropy gain exploration for long horizon multi-goal reinforcement learning.
\newblock In \emph{International Conference on Machine Learning}, pp.\  7750--7761. PMLR, 2020.

\bibitem[Rho et~al.(2024)Rho, Smith, Li, Levine, Peng, and Ha]{rho_language_2024}
Seungeun Rho, Laura Smith, Tianyu Li, Sergey Levine, Xue~Bin Peng, and Sehoon Ha.
\newblock Language {Guided} {Skill} {Discovery}, June 2024.
\newblock URL \url{http://arxiv.org/abs/2406.06615}.
\newblock arXiv:2406.06615 [cs].

\bibitem[Sancaktar et~al.(2022)Sancaktar, Blaes, and Martius]{sancaktar2022curious}
Cansu Sancaktar, Sebastian Blaes, and Georg Martius.
\newblock Curious exploration via structured world models yields zero-shot object manipulation.
\newblock \emph{Advances in Neural Information Processing Systems}, 35:\penalty0 24170--24183, 2022.

\bibitem[Settles(2009)]{settles2009active}
Burr Settles.
\newblock Active learning literature survey, 2009.

\bibitem[Sharma et~al.(2020)Sharma, Gu, Levine, Kumar, and Hausman]{sharma_dynamics-aware_2020}
Archit Sharma, Shixiang Gu, Sergey Levine, Vikash Kumar, and Karol Hausman.
\newblock Dynamics-{Aware} {Unsupervised} {Discovery} of {Skills}, February 2020.
\newblock URL \url{http://arxiv.org/abs/1907.01657}.
\newblock arXiv:1907.01657.

\bibitem[Strouse et~al.(2022)Strouse, Baumli, Warde-Farley, Mnih, and Hansen]{strouse_learning_2022_optimistic}
D.~J. Strouse, Kate Baumli, David Warde-Farley, Vlad Mnih, and Steven Hansen.
\newblock Learning more skills through optimistic exploration, May 2022.
\newblock URL \url{http://arxiv.org/abs/2107.14226}.
\newblock arXiv:2107.14226 [cs].

\bibitem[Sukhija et~al.(2023)Sukhija, Treven, Sancaktar, Blaes, Coros, and Krause]{sukhija_optimistic_2023}
Bhavya Sukhija, Lenart Treven, Cansu Sancaktar, Sebastian Blaes, Stelian Coros, and Andreas Krause.
\newblock Optimistic {Active} {Exploration} of {Dynamical} {Systems}, October 2023.
\newblock URL \url{http://arxiv.org/abs/2306.12371}.
\newblock arXiv:2306.12371 [cs, eess].

\bibitem[Sukhija et~al.(2024)Sukhija, Coros, Krause, Abbeel, and Sferrazza]{sukhija2024maxinforl}
Bhavya Sukhija, Stelian Coros, Andreas Krause, Pieter Abbeel, and Carmelo Sferrazza.
\newblock Maxinforl: Boosting exploration in reinforcement learning through information gain maximization.
\newblock \emph{arXiv preprint arXiv:2412.12098}, 2024.

\bibitem[Tassa et~al.(2018)Tassa, Doron, Muldal, Erez, Li, Casas, Budden, Abdolmaleki, Merel, Lefrancq, et~al.]{tassa2018deepmind}
Yuval Tassa, Yotam Doron, Alistair Muldal, Tom Erez, Yazhe Li, Diego de~Las Casas, David Budden, Abbas Abdolmaleki, Josh Merel, Andrew Lefrancq, et~al.
\newblock Deepmind control suite.
\newblock \emph{arXiv preprint arXiv:1801.00690}, 2018.

\bibitem[Thompson(1933)]{thompson1933likelihood}
William~R Thompson.
\newblock On the likelihood that one unknown probability exceeds another in view of the evidence of two samples.
\newblock \emph{Biometrika}, 25\penalty0 (3-4):\penalty0 285--294, 1933.

\bibitem[Tirinzoni et~al.(2025)Tirinzoni, Touati, Farebrother, Guzek, Kanervisto, Xu, Lazaric, and Pirotta]{tirinzoni2025zeroshot}
Andrea Tirinzoni, Ahmed Touati, Jesse Farebrother, Mateusz Guzek, Anssi Kanervisto, Yingchen Xu, Alessandro Lazaric, and Matteo Pirotta.
\newblock Zero-shot whole-body humanoid control via behavioral foundation models.
\newblock In \emph{The Thirteenth International Conference on Learning Representations}, 2025.

\bibitem[Touati \& Ollivier(2021)Touati and Ollivier]{touati2021learning}
Ahmed Touati and Yann Ollivier.
\newblock Learning one representation to optimize all rewards.
\newblock \emph{Advances in Neural Information Processing Systems}, 34:\penalty0 13--23, 2021.

\bibitem[Touati et~al.(2022)Touati, Rapin, and Ollivier]{touati2022does}
Ahmed Touati, J{\'e}r{\'e}my Rapin, and Yann Ollivier.
\newblock Does zero-shot reinforcement learning exist?
\newblock \emph{arXiv preprint arXiv:2209.14935}, 2022.

\bibitem[Vlastelica et~al.(2021)Vlastelica, Blaes, Pinneri, and Martius]{vlastelica2021risk}
Marin Vlastelica, Sebastian Blaes, Cristina Pinneri, and Georg Martius.
\newblock Risk-averse zero-order trajectory optimization.
\newblock In \emph{5th Annual Conference on Robot Learning}, 2021.

\bibitem[Vlastelica et~al.(2024)Vlastelica, Cheng, Martius, and Kolev]{vlastelicaoffline}
Marin Vlastelica, Jin Cheng, Georg Martius, and Pavel Kolev.
\newblock Offline diversity maximization under imitation constraints.
\newblock In \emph{Reinforcement Learning Conference}, 2024.

\bibitem[Wu et~al.(2018)Wu, Tucker, and Nachum]{wu2018laplacian}
Yifan Wu, George Tucker, and Ofir Nachum.
\newblock The laplacian in rl: Learning representations with efficient approximations.
\newblock \emph{arXiv preprint arXiv:1810.04586}, 2018.

\end{thebibliography}
